\documentclass[dvipsnames]{article} 
\usepackage{ijcai21}

\usepackage{times}
\usepackage{soul}
\usepackage{url}
\usepackage[hidelinks]{hyperref}
\usepackage[utf8]{inputenc}
\usepackage[small]{caption}
\usepackage{graphicx}
\usepackage{amsmath}
\usepackage{amsthm}
\usepackage{booktabs}
\usepackage{algorithm}
\usepackage{algorithmic}
\usepackage[T1]{fontenc}
\usepackage{amsfonts}
\usepackage{nicefrac}
\usepackage{microtype}
\usepackage{lipsum}
\usepackage[inline]{enumitem}
\usepackage{amssymb}
\usepackage{dsfont}
\usepackage{bm}
\usepackage[none]{hyphenat}
\usepackage{array}
\usepackage{makecell}
\usepackage{diagbox}
\usepackage{onimage}
\usepackage{tikz}
\usetikzlibrary{spy}
\usepackage{pgfplots}
\usepackage{multirow}
\usepackage{subcaption}
\usepackage{wrapfig}
\usepackage{xr}
\usepackage{natbib}
\usepackage{xcolor}

\title{PriorityCut: Occlusion-guided Regularization for Warp-based Image Animation}

\author{
Wai Ting Cheung
\And
Gyeongsu Chae
\affiliations
MoneyBrain Inc.
\emails
\{ken, gc\}@moneybrain.ai
}

\begin{document}

\maketitle

\begin{abstract}
	Image animation generates a video of a source image following the motion of a driving video.
	State-of-the-art self-supervised image animation approaches warp the source image according to the motion of the driving video
	and recover the warping artifacts by inpainting.
	These approaches mostly use vanilla convolution for inpainting,
	and vanilla convolution does not distinguish between valid and invalid pixels.
	As a result, visual artifacts are still noticeable after inpainting.
	CutMix is a state-of-the-art regularization strategy that cuts and mixes patches of images and is widely studied in different computer vision tasks.
	Among the remaining computer vision tasks, warp-based image animation is one of the fields that the effects of CutMix have yet to be studied.
	This paper first presents a preliminary study on the effects of CutMix on warp-based image animation.
	We observed in our study that CutMix helps improve only pixel values, but disturbs the spatial relationships between pixels.
	Based on such observation, we propose PriorityCut, a novel augmentation approach that uses the top-$k$ percent occluded pixels of the foreground to regularize warp-based image animation.
	By leveraging the domain knowledge in warp-based image animation,
	PriorityCut significantly reduces the warping artifacts in state-of-the-art warp-based image animation models on diverse datasets.
\end{abstract}

\section{Introduction}

Image animation takes a source image and a driving video as inputs and generates a video of the source image that follows the motion of the driving video.
Traditional image animation requires a reference pose of the animated object such as facial keypoints or edge maps~\citep{fu2019high, ha2019marionette, qian2019make, zhang2019faceswapnet,  otberdout2020dynamic}.
Self-supervised image animation does not require explicit keypoint labels on the objects~\citep{wiles2018x2face, kim2019unsupervised, siarohin2019animating, siarohin2019first, yao2020mesh}.
Warp-based self-supervised image animation warps the source image based on the estimated motion of the driving video and recovers the warping artifacts by inpainting.
The warped image contains regions with a mixture of valid and invalid pixels.
Applying vanilla convolutional filters to recover the warping artifacts in such regions causes ambiguity during training and results in visual artifacts such as color inconsistencies, blurry texture, and noticeable edge responses during testing as shown in the literature~\citep{wiles2018x2face, kim2019unsupervised, siarohin2019animating, siarohin2019first, jeon2020cross, yao2020mesh}.

\begin{figure}[t]
	\centering
	\small
	\setlength\tabcolsep{1pt}
	\begin{tabular}{cccccc}
		\makecell{Source                                                     \\ Image}                                                             &
		\makecell{Driving                                                    \\ Image}                                                                     &
		\makecell{Generated                                                  \\ Image}                                                                   &
		\makecell{Occlusion                                                  \\ Mask}                                                                   &
		\makecell{PriorityCut                                                \\ Mask}                                                                 &
		\makecell{Augmented                                                  \\ Image}                                                                     \\
		\includegraphics[width=0.074\textwidth]{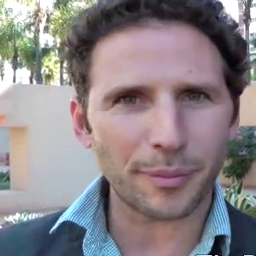}    &
		\includegraphics[width=0.074\textwidth]{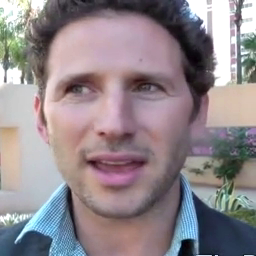}   &
		\includegraphics[width=0.074\textwidth]{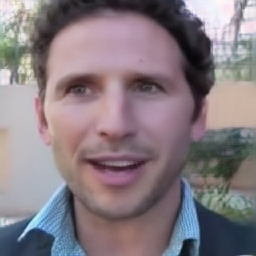} &
		\includegraphics[width=0.074\textwidth]{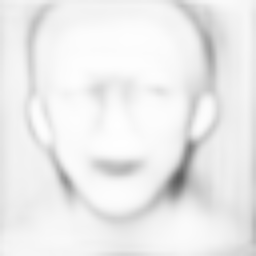} &
		\includegraphics[width=0.074\textwidth]{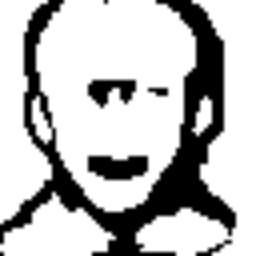}      &
		\includegraphics[width=0.074\textwidth]{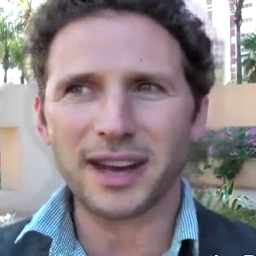}
		\\
	\end{tabular}
	\caption{Warp-based image animation warps the source image based on the motion of the driving image and recovers the warping artifacts by inpainting. PriorityCut utilizes the occlusion information in image animation indicating the locations of warping artifacts to regularize discriminator predictions on inpainting. The augmented image contains a mixture of patches between the driving and the generated images.} 
	\label{fig:pc_front}
\end{figure}


CutMix is a state-of-the-art regularization strategy that cuts and mixes patches of different images to train image classifiers~\citep{yun2019cutmix}.
In addition to image classification, CutMix and its variants are widely studied in different computer vision tasks such as object recognition, object detection, and semantic segmentation~\citep{chen2020gridmask,duta2020pyramidal}.
Most of these tasks deal with only valid pixels in the images,
little is known about the effects of CutMix on a mixture of valid and invalid pixels.
There are recent attempts to apply CutMix on non-warp-based image generation~\citep{schonfeld2020u} and super resolution~\citep{yoo2020rethinking}, which involve a mixture of valid and invalid pixels.
Among the remaining computer vision tasks,
warp-based image animation is one of the fields that the effects of CutMix have yet to be studied.
Our preliminary study on the effects of vanilla CutMix on warp-based image animation indicates that CutMix helps improve only pixel values, but disturbs the spatial relationships between pixels.

\begin{figure}[t]
	\centering
	\small
	\setlength\tabcolsep{3pt}
	\begin{tabular}{cccc}
		Cutout                                                      & Mixup & CutMix & PriorityCut \\
		\includegraphics[width=0.1\textwidth]{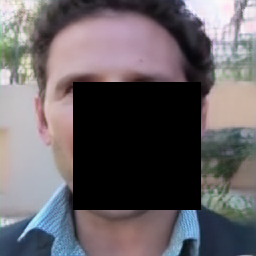} &
		\includegraphics[width=0.1\textwidth]{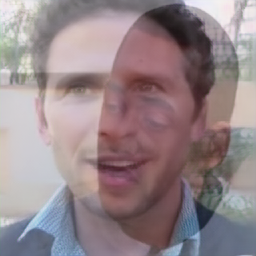}  &
		\includegraphics[width=0.1\textwidth]{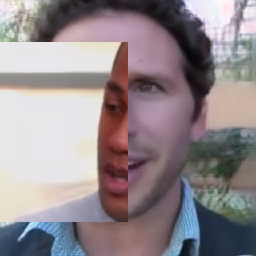} &
		\includegraphics[width=0.1\textwidth]{figures/mask/40/final}                               \\
	\end{tabular}
	\caption{Comparisons of patch-based data augmentation techniques. Cutout~\citep{devries2017improved} drops information. Mixup~\citep{zhang2017mixup} causes a mixture of context. CutMix~\citep{yun2019cutmix} leads to sharp transitions. PriorityCut produces smooth transitions without loss or mixture of context.}
	\label{fig:patch_aug}
\end{figure}

To reduce the ambiguity in recovering the artifacts in image warping, we propose PriorityCut, a novel augmentation inspired by CutMix that uses the top-$k$ percent occluded~pixels of the foreground for consistency regularization.
We observed that the amounts and locations of occlusion are closely related to \textit{how much} and \textit{where} the valid and invalid pixels are.
Leveraging such domain knowledge, PriorityCut derives a new mask from the occlusion mask and the background mask. 
Using the PriorityCut mask, we apply CutMix operation that cuts and mixes patches of different images to regularize discriminator predictions.
Compared to the vanilla rectangular CutMix mask,
PriorityCut mask is flexible in both shapes and locations.
Also, PriorityCut prevents unrealistic patterns and information loss unlike previous approaches~\citep{devries2017improved, yun2019cutmix, zhang2017mixup} shown in Figure~\ref{fig:patch_aug}. 
The subtle differences in our augmented image allow the generator to take small steps in learning,
thus refining the details necessary for realistic inpainting.
We implemented PriorityCut on top of the state-of-the-art image animation model~\citep{siarohin2019first} and evaluated PriorityCut on
\textsf{VoxCeleb}~\citep{nagrani2017voxceleb}, \textsf{BAIR}~\citep{ebert2017self}, and \textsf{Tai-Chi-HD}~\citep{siarohin2019first}
datasets.
Our experimental results show that PriorityCut significantly reduces the visual artifacts in state-of-the-art warp-based image animation models.


%

Our contributions are summarized as follows:
\begin{enumerate}
	\item To the best of our knowledge, we are the first to study the effects of CutMix and its variants on warp-based image animation.
	\item We proposed PriorityCut, a novel augmentation that leverages the domain knowledge of warp-based image animation for regularization.
	\item Our extensive evaluations on diverse datasets with solid baselines show that PriorityCut significantly reduces the warping artifacts in state-of-the-art warp-based image animation models.
\end{enumerate}

\section{Related Work}
\label{sec:related}

\paragraph{Warp-based image animation}
Warp-based image animation warps the source image based
on the estimated motion of the driving video and recovers the
warping artifacts by inpainting.
X2Face~\citep{wiles2018x2face} uses an embedding network and a driving network to generate images.
\cite{kim2019unsupervised} used a keypoint detector and a motion generator to predict videos of an action class based on a single image.
Monkey-Net~\citep{siarohin2019animating} generates images based on a source image, relative keypoint movements, and dense motion.
First~Order~Motion~Model~\citep{siarohin2019first} extended Monkey-Net by predicting Jacobians in keypoint detection and an occlusion~mask.
\citet{yao2020mesh} generated images based on optical flow predicted on 3D meshes.

Warping images produce regions with a mixture of valid and invalid pixels.
Vanilla convolutional layers treat all pixels as valid ones.
When applying vanilla convolutional filters to recover the artifacts of a warped image,
warped, unwarped, and synthesized pixels or features in different layers are all treated equally.
This not only causes ambiguity during training,
but also results in color inconsistencies, blurry texture, and noticeable edge responses during testing.



\paragraph{Applications of patch-based data augmentation in computer vision}
Patch-based data augmentation has been widely studied in the literature.
Cutout and its variants drop random patches of an image~\citep{devries2017improved, singh2018hide, chen2020gridmask}.
Mixup blends two images to generate a new sample~\citep{zhang2017mixup}.
CutMix cuts and mixes patches of random regions between images~\citep{yun2019cutmix}.
Given the success of CutMix, CutMix and its variants have been explored in a wide variety of computer vision tasks such as object recognition, object detection, and semantic segmentation~\citep{chen2020gridmask,duta2020pyramidal}.
Vanilla convolutional layers treat all pixels as valid ones.
Using vanilla convolution makes sense for the tasks above as only real images or videos are involved.
There are also recent attempts to apply CutMix on non-warp-based image generation~\citep{schonfeld2020u} and super resolution~\citep{yoo2020rethinking}, which involve a mixture of valid and invalid pixels.
Among the remaining computer vision tasks,
warp-based image animation is one of the fields that the effects of CutMix have yet to be studied.

\section{Methodology}
\label{sec:methodology}


To understand the motivation of PriorityCut,
we first present our preliminary study on the effects of vanilla CutMix on warp-based image animation.
Then, we discuss the effects of vanilla convolution on regions with a mixture of valid and invalid pixels.
Lastly, we present PriorityCut based on the observations of our preliminary study and vanilla convolution.

\subsection{Preliminary Study}
\label{sec:preliminary}

Given the lack of studies on the effects of CutMix on image animation,
we first directly apply vanilla CutMix to an existing image animation model.
We implemented vanilla CutMix on top of First~Order~Motion~Model (FOMM)~\citep{siarohin2019first},
a state-of-the-art warp-based image animation model.

First Order Motion Model consists of a motion estimation module and an image generation module.
The motion estimation module takes as inputs a source image~\textbf{S} and a driving~image~\textbf{D},
and predicts a dense~motion~field~$\hat{\mathcal{T}}_{\textbf{S} \leftarrow \textbf{D}}$ and an occlusion mask~$\hat{\mathcal{O}}_{\textbf{S} \leftarrow \textbf{D}}$.
The image generation module warps the source image based on the dense motion field~$\hat{\mathcal{T}}_{\textbf{S} \leftarrow \textbf{D}}$ and recovers warping artifacts by inpainting the occluded parts of the source image.

To implement CutMix on top of First Order Motion Model,
we followed~\cite{schonfeld2020u} to extend the discriminator to an U-Net architecture.
The U-Net discriminator~$D^U$ consists of an encoder~$D^U_{enc}$ and a decoder~$D^U_{dec}$ connected by skip connections.
The encoder reuses the same architecture as the original discriminator implemented by its authors.
The decoder mirrors the architecture of the encoder.
The encoder predicts whether an image is real or fake as a whole while the decoder predicts if individual pixels are real or fake.
We applied vanilla CutMix between driving and generated images,
and trained the model with an additional consistency regularization loss as in~\cite{schonfeld2020u}.
The consistency regularization loss regularizes the U-Net decoder output on the CutMix image and the CutMix between the U-Net decoder
outputs on real and fake images.

Table~\ref{tab:preliminary} compares First Order Motion Model with and without CutMix for video reconstruction on \textsf{VoxCeleb} dataset.
The \textsf{VoxCeleb} dataset~\citep{nagrani2017voxceleb} is a face dataset with 22,496 videos from YouTube.
We followed the same protocol of FOMM for video reconstruction and evaluation.
We reconstruct a video by using the first frame as the source image and each of the subsequent frames as the driving image.
We evaluated the reconstructed videos in terms of pixel-wise differences ($\mathcal{L}_1$), motion distances measured by external keypoint detector (AKD), and identity embedding distances measured by external face detector (AED).

The findings of our preliminary study are consistent with those of \cite{schonfeld2020u}:
regularizing the discriminator with CutMix improves the overall realness of the generated images.
Surprisingly, however, both motion~transfer~(AKD) and identity preservation (AED) became worse after regularizing with CutMix.
This indicates that only the realness of \textit{pixels} is improved,
but the \textit{spatial~relationships} between pixels became worse.
Accurate motion transfer and identity preservation require the right pixels at the right locations.
Apparently, vanilla CutMix is not directly applicable to warp-based image animation unlike non-warp-based image generation~\citep{schonfeld2020u}, since motion and identity are not properties to be considered when applying CutMix in ~\cite{schonfeld2020u}.
Instead of helping the model to distinguish between valid and invalid pixels,
vanilla CutMix further adds ambiguity to the image generation process.
To understand why this occurs, we need to look at the inner operations of convolution layers.

\begin{table}[t]
	\def\arraystretch{1.2}
	\setlength\tabcolsep{1.5pt}
	\centering
	\small
	\begin{tabular}{@{\extracolsep{0.005\textwidth}} lccc}
		\toprule
		Architecture & $\mathcal{L}_1$~$\downarrow$              & AKD~$\downarrow$                       & AED~$\downarrow$                        \\
		\midrule
		FOMM U-Net   & 0.0401\tiny{$\pm$9e-5}                    & 1.278\tiny{$\pm$2e-3}                  & 0.1347\tiny{$\pm$6e-4}                  \\
		+ CutMix     & \textcolor{Green}{0.0394}\tiny{$\pm$9e-5} & \textcolor{Red}{1.295}\tiny{$\pm$2e-3} & \textcolor{Red}{0.1365}\tiny{$\pm$6e-4} \\
		\bottomrule
	\end{tabular}
	\caption{Comparisons of First Order Motion Model with and without CutMix for video reconstruction on \textsf{VoxCeleb}. Green and red colors indicate better and worse results, respectively.}
	\label{tab:preliminary}
\end{table}

\subsection{Vanilla Convolution}
\label{sec:conv}

The inner operation of a convolution filter can be expressed as $g(x, y) = \omega \ast f(x, y)$, where
\begin{equation} \label{eq:conv}
	\omega \ast f(x, y) = \sum_{dx=-a}^{a} \sum_{dy=-b}^{b} \omega (dx, dy) f(x + dx, y + dy)
\end{equation}
$g(x, y)$ is the filtered image,
$f(x, y)$ is the original image,
$\omega$~is the filter kernel.
Every element of the filter kernel is considered by
$-a \le dx \le a$ and $-b \le dy \le b$.
For simplicity, the bias term is ignored.

In warp-based image animation,
$f(x, y)$ is the warped image, which contains regions with a mixture of valid and invalid pixels.
Equation~\ref{eq:conv} applies the same convolution filter to both valid and invalid pixels, which leads to ambiguity during training and visual artifacts during testing.
The context becomes even more ambiguous when discriminator provides feedback based on an image augmented by vanilla CutMix.
Consider the CutMix image having sharp transitions in Figure~\ref{fig:patch_aug}.
This helps the discriminator learn only the realness of pixels as in~\cite{schonfeld2020u},
but not the spatial relationships.
This explains the observation in our preliminary study that vanilla CutMix improves only the pixel values~($\mathcal{L}_1$) in warp-based image animation,
but worsen motion transfer (AKD) and identity preservation (AED).
To reduce the ambiguity,
we need a CutMix operation that follows the context of warp-based image animation.

\subsection{PriorityCut}
\label{sec:priority-cut}
To capture the context of warp-based image animation,
PriorityCut leverages the occlusion information to perform CutMix augmentation.
Our approach is based on two key observations.
One observation is that occlusion in warping-based image animation reflects the degrees of validity of the visual artifacts.
Another observation is that it contains the spatial relations between regions of valid and invalid pixels.
The occlusion information can serve as a guidance to the vanilla convolution layers of discriminator \textit{how much} and \textit{where} the valid and invalid pixels are.

\begin{figure}[!tbp]
	\centering
	\small
	\def\arraystretch{1}
	\setlength\tabcolsep{1pt}
	\begin{tabular}{*7{>{\centering\arraybackslash} c}}
		Source                                                            &
		Driving                                                           &
		Gen                                                               &
		\makecell{Occ                                                       \\ Mask}                                                        &
		\makecell{Bg                                                        \\ Mask}                                                       &
		\makecell{PriorityCut                                               \\ Mask}                                                      &
		Aug
		\\
		\includegraphics[width=0.06\textwidth]{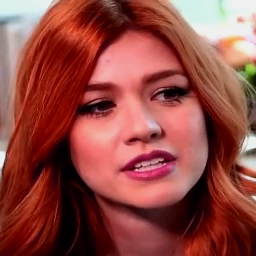}    &
		\includegraphics[width=0.06\textwidth]{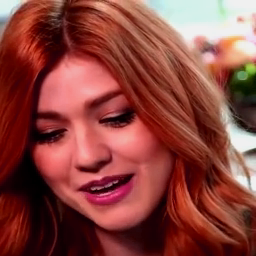}   &
		\includegraphics[width=0.06\textwidth]{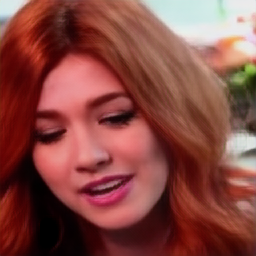} &
		\includegraphics[width=0.06\textwidth]{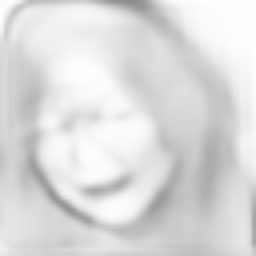} &
		\includegraphics[width=0.06\textwidth]{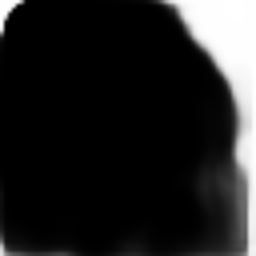}      &
		\includegraphics[width=0.06\textwidth]{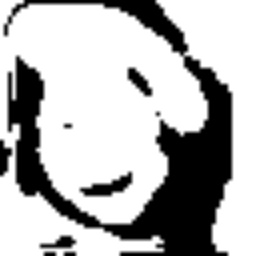}      &
		\includegraphics[width=0.06\textwidth]{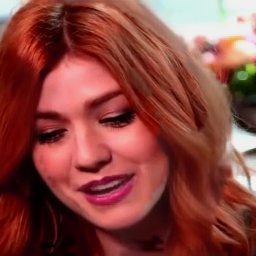}
		\\
	\end{tabular}
	\caption{Illustration of deriving PriorityCut masks from occlusion and background masks based on the top 30\% occlusion ($k=30$).}
	\label{fig:mask}
\end{figure}

Based on the above observations, we propose PriorityCut, a novel augmentation that uses the top-$k$~percent occluded pixels of the foreground as the CutMix mask.
Figure~\ref{fig:mask} illustrates the derivation of PriorityCut masks from occlusion and background masks predicted by a warp-based image animation model.
Suppose $\mathcal{M}_{bg}$ is a predicted alpha background mask, ranging between 0~and~1.
We first suppress the uncertain pixels of the alpha~background~mask~$\mathcal{M}_{bg}$ to obtain a binary~background~mask~$\hat{\mathcal{M}}_{bg}$.
$\hat{\mathcal{M}}_{bg}$~corresponds to the background~mask predicted with high confidence.
The occlusion map~$\hat{\mathcal{O}}_{\textbf{S} \leftarrow \textbf{D}} \in [0, 1]^{H \times W}$ is an alpha mask, with 0 being fully occluded and 1 being not occluded.
Equation~\ref{eq:occ} utilizes $\hat{\mathcal{M}}_{bg}$ to compute the occlusion map of the foreground~$\hat{\mathcal{O}}_{fg}$:
\begin{equation} \label{eq:occ}
	\hat{\mathcal{O}}_{fg} = \hat{\mathcal{M}}_{bg} + (1 - \hat{\mathcal{M}}_{bg}) \odot \hat{\mathcal{O}}_{\textbf{S} \leftarrow \textbf{D}}
\end{equation}
where $\odot$ denotes the Hadamard product.
It retains only the foreground portions of the occlusion masks shown in Figure~\ref{fig:mask}, which are also alpha masks.
Given a percentile~$k$, we denote the PriorityCut~mask~$\mathcal{M}_{pc}$
as the top-$k$~percent occluded~pixels of the foreground~$\hat{\mathcal{O}}_{fg}$.
Following~\cite{yun2019cutmix}, we randomize the values of $k$ in our experiments.
Equation~\ref{eq:cutmix} utilizes the PriorityCut mask to perform CutMix between the real images~$x$ and the generated images~$x'$.
To avoid sharp transitions, PriorityCut performs CutMix on the driving image~\textbf{D} and its reconstruction~$\hat{\textbf{D}}$.
\begin{equation} \label{eq:cutmix}
	\text{mix}(x, x', \mathcal{M}_{pc}) = \mathcal{M}_{pc} \odot x + (1 - \mathcal{M}_{pc}) \odot x'
\end{equation}


\begin{figure}[!tbp]
	\centering
	\small
	\setlength\tabcolsep{1.5pt}
	\begin{tabular}{cccccc}
		Driving                                                               & \makecell{Vanilla CutMix \\ Predictions} & \makecell{PriorityCut \\ Predictions} \\
		\includegraphics[width=0.075\textwidth]{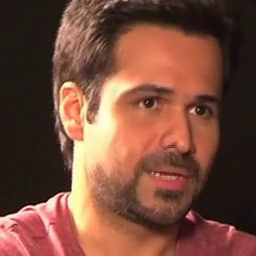}     &
		\includegraphics[width=0.075\textwidth]{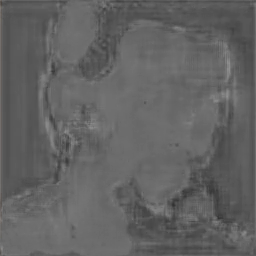} &
		\includegraphics[width=0.075\textwidth]{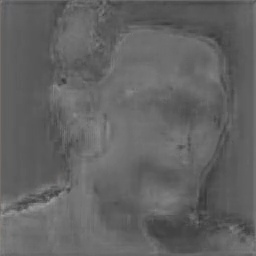}                            \\
	\end{tabular}
	\caption{Comparison of discriminator predictions on individual pixels between vanilla CutMix and PriorityCut.}
	\label{fig:feedback}
\end{figure}


In Figure~\ref{fig:mask}, the augmented image looks almost identical to the driving or the generated images with only subtle differences in fine details.
PriorityCut always assigns the fake pixels to locations where there are large changes in motion,
creating incentives for the generator to improve.
For example, borders, edges, in-between regions of distinct objects (e.g. face, mic, wall), or parts of objects (e.g. hair, eyes, nose, mouth).
The design philosophy of PriorityCut follows that of CutBlur~\citep{yoo2020rethinking}.
The augmented images have no sharp transitions, mixed image contents, or loss of the relationships of pixels unlike previous approaches shown in Figure~\ref{fig:patch_aug}.
PriorityCut also adds another degree of flexibility to the mask shapes.
The discriminator can no longer rely on a rectangular area like the vanilla CutMix to predict where the real and fake pixels concentrate at.
This encourages the discriminator to learn properly the locations of the real and fake pixels.
Figure~\ref{fig:feedback} compares the discriminator predictions on individual pixels between PriorityCut and vanilla CutMix.
PriorityCut helps the discriminator learn clear distinctions between real and fake pixels around locations with large changes in motion.
In contrast, vanilla CutMix helps the discriminator learn only vague estimations.
This supports our observations in Sections~\ref{sec:preliminary} and~\ref{sec:conv} that vanilla CutMix is not good at learning spatial relationships.

\section{Experiments}
\label{sec:experiments}

\subsection{Experimental setup}
\label{sec:expr_setup}



\paragraph{Datasets}
We followed~\citep{siarohin2019first} to preprocess high-quality videos on the following datasets and resized them to $256 \times 256$ resolution:
the~\textsf{VoxCeleb}~dataset~\citep{nagrani2017voxceleb} (18,398 training and 512 testing videos after preprocessing);
the \textsf{Tai-Chi-HD} dataset~\citep{siarohin2019first} (2,994 training and 285 testing video chunks after preprocessing);
the \textsf{BAIR} robot pushing dataset~\citep{ebert2017self} (42,880~training and 128~testing~videos).

\paragraph{Evaluation protocol}
We followed \citep{siarohin2019first} to quantitatively and qualitatively evaluate video reconstruction.
For video reconstruction, we used the first frame of the input video as the source image and each frame as the driving image.
We evaluated the reconstructed videos against the ground~truth videos on the following metrics:
\textit{pixel-wise differences} ($\mathcal{L}_1$); \textit{PSNR}, \textit{SSIM}, and their masked versions (\textit{M-PSNR}, \textit{M-SSIM}); \textit{average keypoint distance (AKD)}, \textit{missing keypoint rate~(MKR)}, and \textit{average~Euclidean~distance~(AED)} of feature embeddings detected by third-party tools.
For details on dataset preprocessing and metric computation, refer to Section~\ref{sec:appx_eval} in the appendix.

\subsection{Comparison with state-of-the-art}

We quantitatively and qualitatively compared PriorityCut with state-of-the-art warp-based image animation methods with publicly available implementations.

\begin{itemize}[leftmargin=*]
	\item \textbf{X2Face}. The reenactment system with an embedding and a driving network~\citep{wiles2018x2face}.

	\item \textbf{Monkey-Net}. The motion transfer framework based on a keypoint detector, a dense motion network, and a motion transfer generator~\citep{siarohin2019animating}.

	\item \textbf{First Order Motion Model}. The motion transfer network that extends Monkey-Net by estimating affine transformations for the keypoints and predicting occlusion for inpainting~\citep{siarohin2019first}.
	      We compared two versions of First~Order~Motion~Model.
	      The baseline model~(FOMM) corresponds to the one in their published paper.
	      The adversarial model~(FOMM+) is a concurrent work with an adversarial discriminator.
	      Since its authors have released\footnote{\url{https://github.com/AliaksandrSiarohin/first-order-model}} both models,
	      we evaluated the baseline model and additionally the adversarial model.

	\item \textbf{Ours}. Our implementation using First Order Motion Model as backbone with U-Net discriminator to provide discriminator predictions on individual pixels and PriorityCut to regularize inpainting.
\end{itemize}

\begin{table*}[!tbp]
	\def\arraystretch{1.2}
	\setlength\tabcolsep{1.5pt}
	\centering
	\small
	\begin{tabular}{@{\extracolsep{0.005\textwidth}}*{10}{l}}
		\toprule
		Model      & \multicolumn{1}{c}{$\mathcal{L}_1$~$\downarrow$} & \multicolumn{3}{c}{PSNR~$\uparrow$}    & \multicolumn{3}{c}{SSIM~$\uparrow$}    & \multicolumn{1}{c}{AKD~$\downarrow$} & \multicolumn{1}{c}{AED~$\downarrow$}                                                                                                                                                                    \\
		\cmidrule{3-5} \cmidrule{6-8}
		           &                                                  & \multicolumn{1}{c}{All}                & \multicolumn{1}{c}{Salient}            & \multicolumn{1}{c}{$\neg$ Salient}   & \multicolumn{1}{c}{All}                & \multicolumn{1}{c}{Salient}            & \multicolumn{1}{c}{$\neg$ Salient} &                                        &                                         \\
		\midrule
		X2Face     & 0.0739\tiny{$\pm$2e-4}                           & 19.13\tiny{$\pm$0.02}                  & 20.04\tiny{$\pm$0.02}                  & 30.65\tiny{$\pm$0.04}                & 0.625\tiny{$\pm$6e-4}                  & 0.681\tiny{$\pm$5e-4}                  & 0.944\tiny{$\pm$2e-4}              & 6.847\tiny{$\pm$4e-3}                  & 0.3664\tiny{$\pm$2e-3}                  \\
		Monkey-Net & 0.0477\tiny{$\pm$1e-4}                           & 22.47\tiny{$\pm$0.02}                  & 23.29\tiny{$\pm$0.02}                  & 34.43\tiny{$\pm$0.04}                & 0.730\tiny{$\pm$5e-4}                  & 0.769\tiny{$\pm$4e-4}                  & 0.962\tiny{$\pm$2e-4}              & 1.892\tiny{$\pm$4e-3}                  & 0.1967\tiny{$\pm$8e-4}                  \\
		FOMM       & 0.0413\tiny{$\pm$9e-5}                           & \underline{24.28}\tiny{$\pm$0.02}      & \underline{25.19}\tiny{$\pm$0.02}      & 36.19\tiny{$\pm$0.04}                & \underline{0.791}\tiny{$\pm$4e-4}      & \underline{0.825}\tiny{$\pm$4e-4}      & \underline{0.969}\tiny{$\pm$2e-4}  & \underline{1.290}\tiny{$\pm$2e-3}      & \underline{0.1324}\tiny{$\pm$6e-4}      \\
		FOMM+      & \underline{0.0409}\tiny{$\pm$9e-5}               & \textcolor{Red}{24.26}\tiny{$\pm$0.02} & \textcolor{Red}{25.17}\tiny{$\pm$0.02} & \underline{36.26}\tiny{$\pm$0.04}    & \textcolor{Red}{0.790}\tiny{$\pm$4e-4} & \textcolor{Red}{0.822}\tiny{$\pm$4e-4} & \textbf{0.970}\tiny{$\pm$1e-4}     & \textcolor{Red}{1.305}\tiny{$\pm$2e-3} & \textcolor{Red}{0.1339}\tiny{$\pm$6e-4} \\
		Ours       & \textbf{0.0401}\tiny{$\pm$9e-5}                  & \textbf{24.45}\tiny{$\pm$0.02}         & \textbf{25.35}\tiny{$\pm$0.02}         & \textbf{36.45}\tiny{$\pm$0.04}       & \textbf{0.793}\tiny{$\pm$4e-4}         & \textbf{0.826}\tiny{$\pm$2e-4}         & \textbf{0.970}\tiny{$\pm$1e-4}     & \textbf{1.286}\tiny{$\pm$2e-3}         & \textbf{0.1303}\tiny{$\pm$6e-4}         \\
		\bottomrule
	\end{tabular}
	\caption{Comparison with state-of-the-art approaches for video reconstruction on \textsf{VoxCeleb}. Bold and underline indicate the best and the second best results, respectively. For variants of FOMM that do not produce the best or the second best results, red indicates worse results compared to the baseline FOMM.}
	\label{tab:sota_vox}
\end{table*}

\begin{table*}[!tbp]
	\def\arraystretch{1.2}
	\setlength\tabcolsep{1.5pt}
	\centering
	\small
	\begin{tabular}{l @{\extracolsep{0.005\textwidth}}*{10}{r}}
		\toprule
		Model
		           &
		\multicolumn{1}{c}{$\mathcal{L}_1$~$\downarrow$}
		           &
		\multicolumn{3}{c}{PSNR~$\uparrow$}
		           &
		\multicolumn{3}{c}{SSIM~$\uparrow$}
		           &
		\multicolumn{1}{c}{AKD~$\downarrow$}
		           &
		\multicolumn{1}{c}{MKR~$\downarrow$}
		           &
		\multicolumn{1}{c}{AED~$\downarrow$}                                                                                                                                                                                                                                                                                                                                                  \\
		\cmidrule{3-5} \cmidrule{6-8}
		           &                                    & \multicolumn{1}{c}{All}           & \multicolumn{1}{c}{Salient}       & \multicolumn{1}{c}{$\neg$ Salient} & \multicolumn{1}{c}{All}           & \multicolumn{1}{c}{Salient}       & \multicolumn{1}{c}{$\neg$ Salient} &                                  &                                                                        \\
		\midrule
		X2Face     & 0.0729\tiny{$\pm$3e-4}             & 18.16\tiny{$\pm$0.02}             & 21.08\tiny{$\pm$0.02}             & 22.24\tiny{$\pm$0.02}              & 0.580\tiny{$\pm$1e-3}             & 0.858\tiny{$\pm$3e-4}             & 0.734\tiny{$\pm$1e-3}              & 14.89\tiny{$\pm$8e-2}            & 0.175\tiny{$\pm$1e-3}             & 0.2441\tiny{$\pm$6e-4}             \\
		Monkey-Net & 0.0691\tiny{$\pm$3e-4}             & 18.89\tiny{$\pm$0.03}             & 22.02\tiny{$\pm$0.03}             & 22.70\tiny{$\pm$0.04}              & 0.599\tiny{$\pm$2e-3}             & 0.867\tiny{$\pm$3e-4}             & 0.742\tiny{$\pm$1e-3}              & 11.40\tiny{$\pm$7e-2}            & 0.060\tiny{$\pm$7e-4}             & 0.2319\tiny{$\pm$7e-4}             \\
		FOMM       & 0.0569\tiny{$\pm$2e-4}             & 21.29\tiny{$\pm$0.03}             & 24.65\tiny{$\pm$0.03}             & 25.18\tiny{$\pm$0.04}              & 0.651\tiny{$\pm$2e-3}             & 0.891\tiny{$\pm$3e-4}             & \underline{0.771}\tiny{$\pm$1e-3}  & 6.87\tiny{$\pm$6e-2}             & 0.038\tiny{$\pm$5e-4}             & 0.1657\tiny{$\pm$6e-4}             \\
		FOMM+      & \underline{0.0555}\tiny{$\pm$2e-4} & \underline{21.35}\tiny{$\pm$0.03} & \underline{24.74}\tiny{$\pm$0.03} & \underline{25.21}\tiny{$\pm$0.04}  & \textbf{0.654}\tiny{$\pm$2e-3}    & \underline{0.893}\tiny{$\pm$3e-4} & \textbf{0.772}\tiny{$\pm$1e-3}     & \textbf{6.73}\tiny{$\pm$6e-2}    & \underline{0.032}\tiny{$\pm$4e-4} & \underline{0.1647}\tiny{$\pm$6e-4} \\
		Ours       & \textbf{0.0549}\tiny{$\pm$2e-4}    & \textbf{21.54}\tiny{$\pm$0.03}    & \textbf{24.98}\tiny{$\pm$0.03}    & \textbf{25.33}\tiny{$\pm$0.04}     & \underline{0.653}\tiny{$\pm$2e-3} & \textbf{0.896}\tiny{$\pm$3e-4}    & 0.768\tiny{$\pm$1e-3}              & \underline{6.78}\tiny{$\pm$6e-2} & \textbf{0.030}\tiny{$\pm$4e-4}    & \textbf{0.1629}\tiny{$\pm$6e-4}    \\
		\bottomrule
	\end{tabular}
	\caption{Comparison with state-of-the-art approaches for video reconstruction on \textsf{Tai-Chi-HD}. Bold and underline indicate the best and the second best results, respectively.}
	\label{tab:sota_taichi}
\end{table*}

\paragraph{Quantitative comparison}
Table~\ref{tab:sota_vox},~\ref{tab:sota_taichi},~and~\ref{tab:sota_bair} show the quantitative comparison results of video reconstruction on the \textsf{VoxCeleb}, \textsf{BAIR}, and \textsf{Tai-Chi-HD} datasets, respectively.
For all tables, the down arrows indicate that lower values mean better results, and vice versa.
We show the 95\%~confidence intervals,
highlight the best results in bold and underline the second-best.
For variants of the baseline model that do not produce the best or the second best results,
the red and green texts indicate worse and better results than the baseline, respectively.
This serves a similar purpose as the ablation study, indicating the effectiveness of certain components in improving the baseline.
Since the pre-trained model of FOMM+ for \textsf{BAIR} is not publicly available at the time of writing,
we evaluated \textsf{BAIR} only on the baseline FOMM.

PriorityCut outperforms state-of-the-art models in every single metric for \textsf{VoxCeleb} and \textsf{BAIR},
and in most of the metrics for \textsf{Tai-Chi-HD}.
Note that adversarial training alone (FOMM+) does not always guarantee improvements, as highlighted in red for \textsf{VoxCeleb}.
\begin{table}
	\def\arraystretch{1.2}
	\setlength\tabcolsep{3pt}
	\centering
	\small
	\begin{tabular}{@{\extracolsep{\fill}}*{4}{l}}
		\toprule
		Model      & \multicolumn{1}{c}{$\mathcal{L}_1$~$\downarrow$} & \multicolumn{1}{c}{PSNR~$\uparrow$}   & \multicolumn{1}{c}{SSIM~$\uparrow$}     \\
		\midrule
		X2Face     & 0.0419\scriptsize{$\pm$5e-4}                     & 21.3\scriptsize{$\pm$0.1}             & 0.831\scriptsize{$\pm$2e-3}             \\
		Monkey-Net & 0.0340\scriptsize{$\pm$4e-4}                     & 23.1\scriptsize{$\pm$0.1}             & 0.867\scriptsize{$\pm$2e-3}             \\
		FOMM       & \underline{0.0292}\scriptsize{$\pm$4e-4}         & \underline{24.8}\scriptsize{$\pm$0.1} & \underline{0.889}\scriptsize{$\pm$1e-3} \\
		Ours       & \textbf{0.0276}\scriptsize{$\pm$3e-4}            & \textbf{25.3}\scriptsize{$\pm$0.1}    & \textbf{0.894}\scriptsize{$\pm$1e-3}    \\
		\bottomrule
	\end{tabular}
	\caption{Comparison with state-of-the-art approaches for video reconstruction on \textsf{BAIR}. Bold and underline indicate the best and the second best results, respectively.}
	\label{tab:sota_bair}
\end{table}

\begin{figure}[t]
	\small
	\setlength\tabcolsep{1.2pt}
	\begin{tabular}{*6{>{\centering\arraybackslash} m{0.075\textwidth}}}
		Source                                                                           & Driving 1 & Driving 2 & Driving 3 & Driving 4 & Driving 5 \\
		\includegraphics[width=0.075\textwidth]{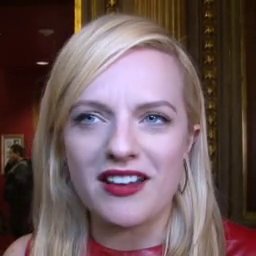}                &
		\includegraphics[width=0.075\textwidth]{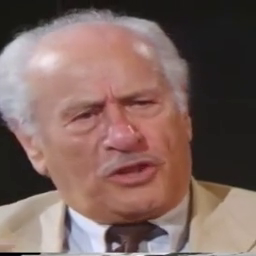}         &
		\includegraphics[width=0.075\textwidth]{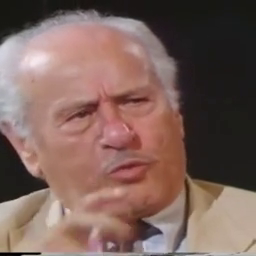}         &
		\includegraphics[width=0.075\textwidth]{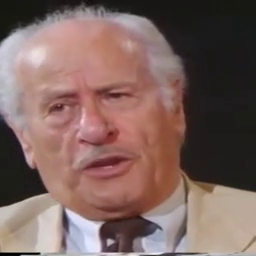}         &
		\includegraphics[width=0.075\textwidth]{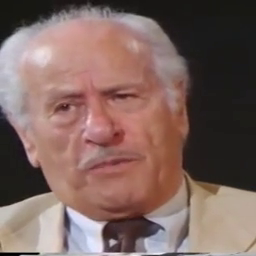}         &
		\includegraphics[width=0.075\textwidth]{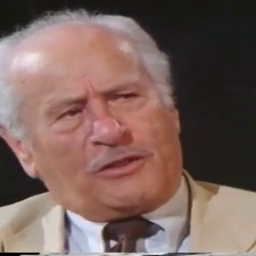}                                                                     \\
		X2Face                                                                           &
		\includegraphics[width=0.075\textwidth]{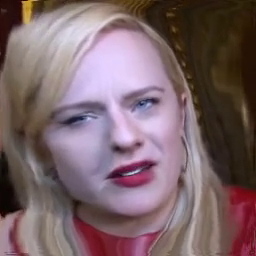}   &
		\includegraphics[width=0.075\textwidth]{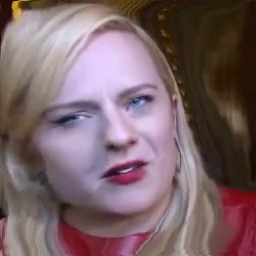}   &
		\includegraphics[width=0.075\textwidth]{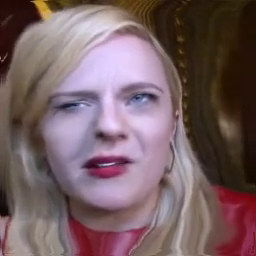}   &
		\includegraphics[width=0.075\textwidth]{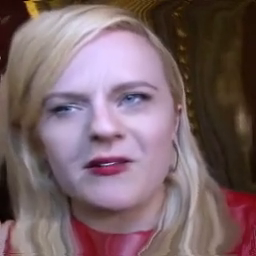}   &
		\includegraphics[width=0.075\textwidth]{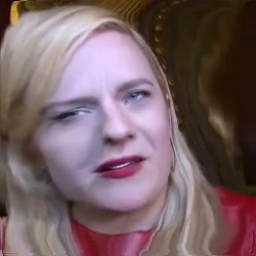}
		\\
		\makecell{Monkey-                                                                                                                            \\ Net}                                                                       &
		\includegraphics[width=0.075\textwidth]{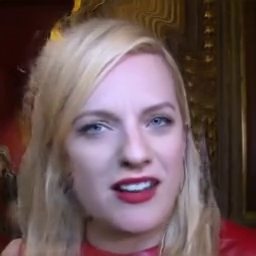}   &
		\includegraphics[width=0.075\textwidth]{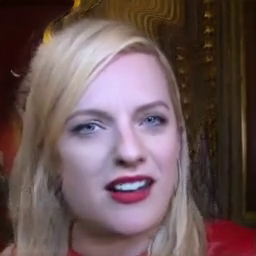}   &
		\includegraphics[width=0.075\textwidth]{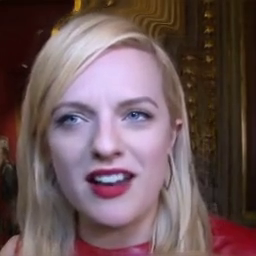}   &
		\includegraphics[width=0.075\textwidth]{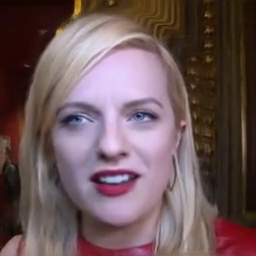}   &
		\includegraphics[width=0.075\textwidth]{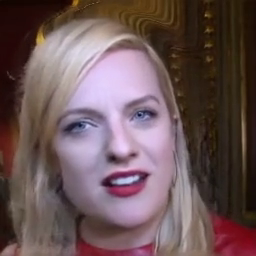}
		\\
		FOMM                                                                             &
		\includegraphics[width=0.075\textwidth]{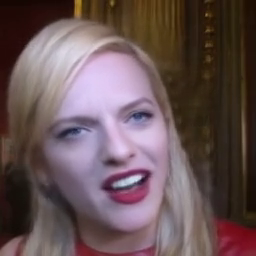} &
		\includegraphics[width=0.075\textwidth]{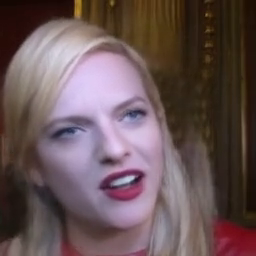} &
		\includegraphics[width=0.075\textwidth]{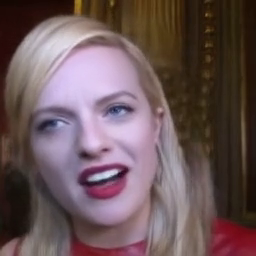} &
		\includegraphics[width=0.075\textwidth]{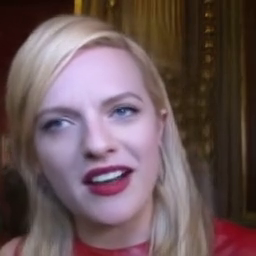} &
		\includegraphics[width=0.075\textwidth]{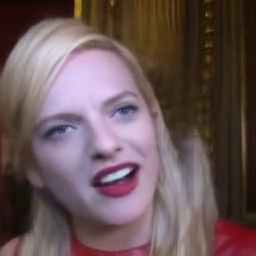}
		\\
		FOMM+                                                                            &
		\includegraphics[width=0.075\textwidth]{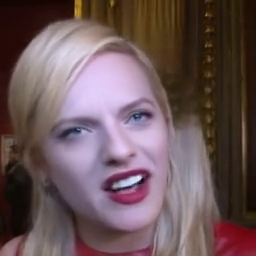}      &
		\includegraphics[width=0.075\textwidth]{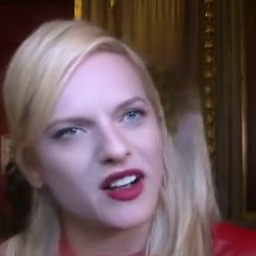}      &
		\includegraphics[width=0.075\textwidth]{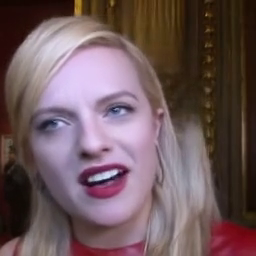}      &
		\includegraphics[width=0.075\textwidth]{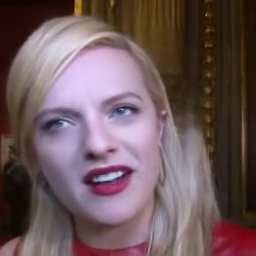}      &
		\includegraphics[width=0.075\textwidth]{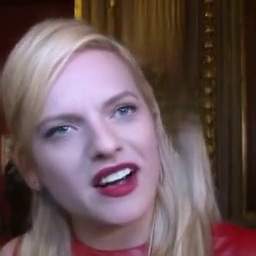}
		\\
		Ours                                                                             &
		\includegraphics[width=0.075\textwidth]{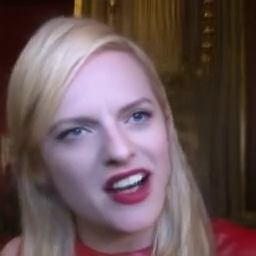} &
		\includegraphics[width=0.075\textwidth]{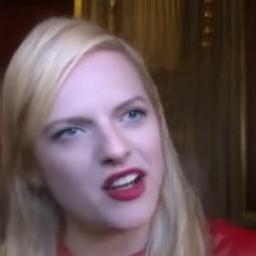} &
		\includegraphics[width=0.075\textwidth]{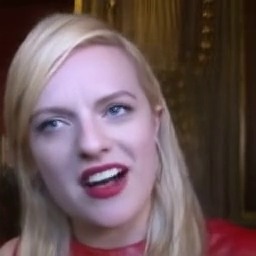} &
		\includegraphics[width=0.075\textwidth]{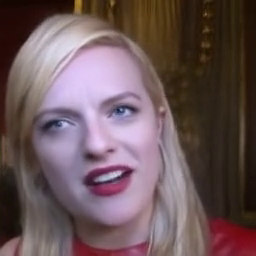} &
		\includegraphics[width=0.075\textwidth]{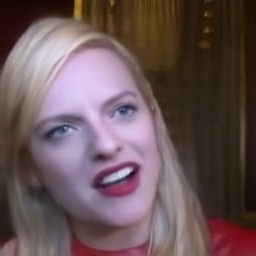}
		\\
	\end{tabular}
	\caption{Qualitative comparison of state-of-the-art approaches for image animation on \textsf{VoxCeleb}.}
	\label{fig:sota}
\end{figure}

\paragraph{Qualitative comparison}
Figure~\ref{fig:sota} shows the qualitative comparison for the \textsf{VoxCeleb} dataset.
X2Face causes identity leak: the model tries to fit the source image into the shapes of the driving images, causing face distortions.
Monkey-Net shows trivial warping artifacts near the right shoulder and on top of the head, and does not closely follow the pose angles.
Since FOMM, FOMM+, and PriorityCut all use the same backbone,
they share similar coarse shapes and geometries and differ in fine details.
First, both FOMM and FOMM+ are weak at spatial relationships: they are aware that some hairs should be inpainted, but the hairs are inpainted at wrong locations.
Also, the baseline FOMM is inconsistent with the amounts of hairs to be inpainted for similar pose angles.
With adversarial training, FOMM+ amplifies the inpainted hairs at wrong locations.
In contrast, PriorityCut inpaints the hairs with proper amounts and locations.
Our qualitative comparison suggests that PriorityCut helps reduce the \textit{ambiguity in spatial relationships} observed in the vanilla convolutional layers of FOMM and FOMM+.
For additional qualitative comparisons, refer to Section~\ref{sec:add_eval} of the appendix and the corresponding video for each figure.

\begin{table*}[!tbp]
	\def\arraystretch{1.2}
	\setlength\tabcolsep{1.5pt}
	\centering
	\small
	\begin{tabular}{@{\extracolsep{0.005\textwidth}} lccccccccc}
		\toprule
		Architecture  & $\mathcal{L}_1$~$\downarrow$              & \multicolumn{3}{c}{PSNR~$\uparrow$}      & \multicolumn{3}{c}{SSIM~$\uparrow$}      & AKD~$\downarrow$                         & AED~$\downarrow$                                                                                                                                                                                        \\
		\cmidrule{3-5} \cmidrule{6-8}
		              &                                           & \multicolumn{1}{c}{All}                  & \multicolumn{1}{c}{Salient}              & \multicolumn{1}{c}{$\neg$ Salient}       & \multicolumn{1}{c}{All}                & \multicolumn{1}{c}{Salient}            & \multicolumn{1}{c}{$\neg$ Salient} &                                        &                                         \\
		\midrule
		Baseline      & 0.0413\tiny{$\pm$9e-5}                    & 24.28\tiny{$\pm$0.02}                    & 25.19\tiny{$\pm$0.02}                    & 36.19\tiny{$\pm$0.04}                    & 0.791\tiny{$\pm$4e-4}                  & \underline{0.825}\tiny{$\pm$4e-4}      & \underline{0.969}\tiny{$\pm$2e-4}  & 1.290\tiny{$\pm$2e-3}                  & \underline{0.1324}\tiny{$\pm$6e-4}      \\
		+ Adv         & \textcolor{Green}{0.0409}\tiny{$\pm$9e-5} & \textcolor{Red}{24.26}\tiny{$\pm$0.02}   & \textcolor{Red}{25.17}\tiny{$\pm$0.02}   & \textcolor{Green}{36.26}\tiny{$\pm$0.04} & \textcolor{Red}{0.790}\tiny{$\pm$4e-4} & \textcolor{Red}{0.822}\tiny{$\pm$2e-4} & \textbf{0.970}\tiny{$\pm$1e-4}     & \textcolor{Red}{1.305}\tiny{$\pm$2e-3} & \textcolor{Red}{0.1339}\tiny{$\pm$6e-4} \\
		+ U-Net       & \underline{0.0401}\tiny{$\pm$9e-5}        & \textcolor{Green}{24.34}\tiny{$\pm$0.02} & \textcolor{Green}{25.29}\tiny{$\pm$0.02} & \textcolor{Green}{36.31}\tiny{$\pm$0.04} & 0.791\tiny{$\pm$4e-4}                  & \textcolor{Red}{0.824}\tiny{$\pm$4e-4} & \underline{0.969}\tiny{$\pm$2e-4}  & \textbf{1.278}\tiny{$\pm$2e-3}         & \textcolor{Red}{0.1347}\tiny{$\pm$6e-4} \\
		+ PriorityCut & \underline{0.0401}\tiny{$\pm$9e-5}        & \underline{24.45}\tiny{$\pm$0.02}        & \underline{25.35}\tiny{$\pm$0.02}        & \textbf{36.45}\tiny{$\pm$0.04}           & \textbf{0.793}\tiny{$\pm$4e-4}         & \textbf{0.826}\tiny{$\pm$4e-4}         & \textbf{0.970}\tiny{$\pm$1e-4}     & \underline{1.286}\tiny{$\pm$2e-3}      & \textbf{0.1303}\tiny{$\pm$6e-4}         \\
		\bottomrule
	\end{tabular}
	\caption{Quantitative ablation study for video reconstruction on \textsf{VoxCeleb}. Bold and underline indicate the best and the second best results, respectively. For results that are not the best or the second best, red and green indicate worse and better results compared to the baseline FOMM, respectively.}
	\label{tab:ablation}
\end{table*}

\subsection{Ablation study}
\label{sec:ablation}

To validate the effects of each proposed component,
we evaluated the following variants of our model. 
\textit{Baseline}: the published First~Order~Motion~Model used in their paper;
\textit{Adv}:~the concurrent work of First~Order~Motion~Model with a global discriminator;
\textit{U-Net}: the architecture of the global discriminator extended to the U-Net architecture; 
\textit{PriorityCut}: our proposed approach that uses the top-$k$ percent occluded pixels of the foreground as the CutMix mask. 

\paragraph{Quantitative ablation study}
Table~\ref{tab:ablation} quantitatively compares the results of video reconstruction on the \textsf{VoxCeleb} dataset~\citep{nagrani2017voxceleb}.
First, adversarial training improves only the $\mathcal{L}_1$ distance and the non-salient parts,
but worsens other metrics.
U-Net discriminator improves $\mathcal{L}_1$ by a margin with better \textit{AKD} as a positive side bonus,
at the cost of further degraded \textit{AED}.
We experimented with adding either PriorityCut or vanilla CutMix on top of the U-Net architecture.
After adding PriorityCut, the full model outperforms the baseline model in every single metric.
In particular, the improvement of \textit{AED} shows the effectiveness of PriorityCut in guiding the model to inpaint realistic features.


\begin{figure}[t]
	\small
	\setlength\tabcolsep{1.2pt}
	\begin{tabular}{*6{>{\centering\arraybackslash} m{0.075\textwidth}}}
		Source                                                                   & Driving & FOMM & +Adv & +U-Net & +PriorityCut \\
		\includegraphics[width=0.075\textwidth]{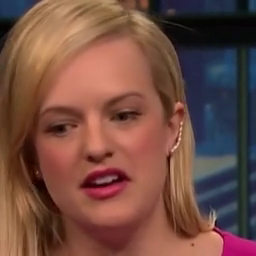}         &
		\includegraphics[width=0.075\textwidth]{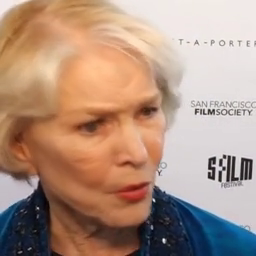}  &
		\includegraphics[width=0.075\textwidth]{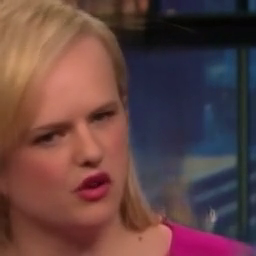} &
		\includegraphics[width=0.075\textwidth]{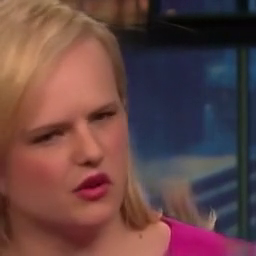}      &
		\includegraphics[width=0.075\textwidth]{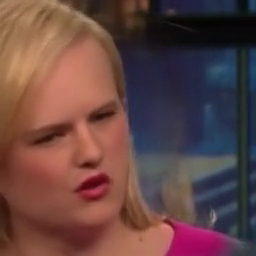}     &
		\includegraphics[width=0.075\textwidth]{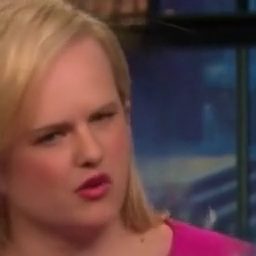}                                                 \\
	\end{tabular}
	\caption{Qualitative ablation study on \textsf{VoxCeleb}.}
	\label{fig:ablation}
\end{figure}

\paragraph{Qualitative ablation study}
Figure~\ref{fig:ablation} qualitatively compares different variants of implementations on \textsf{VoxCeleb} dataset~\citep{nagrani2017voxceleb}.
Similar to Figure~\ref{fig:sota}, the baseline FOMM shows hair artifacts near the shoulder and adversarial training amplifies the artifacts.
Extending the discriminator to U-Net architecture slightly reduces the artifacts.
Finally, adding PriorityCut minimizes the artifacts produced.

\section{Discussion}
\label{sec:discussion}

This section summarizes the key observations and findings.

\paragraph{Limitations of warp-based image animation}

Existing warp-based image animation techniques recover the warping artifacts by inpainting.
We observed that vanilla convolution does not distinguish valid and invalid pixels during inpainting.
Without proper guidance,
the generator struggles at recovering the visual artifacts with proper texture.
While the literature demonstrated the effectiveness of vanilla CutMix in different computer vision tasks,
we observed in our preliminary study that directly applying it to warp-based image animation improves only the pixel values,
ignoring the spatial relationships.

\paragraph{The role of PriorityCut in recovering artifacts}
Based on the above observations,
we proposed PriorityCut to regularize image animation based on domain knowledge from occlusion,
guiding the generator to learn the spatial relationships between pixels.
Our quantitative and qualitative results with solid baselines and diverse datasets show that PriorityCut outperforms state-of-the-art models in recovering the warping artifacts.
Since we implemented PriorityCut on top of FOMM, the artifacts seen in FOMM also exist in our results.
However, PriorityCut alleviates them by reducing the ambiguity of the baseline FOMM.
In addition, PriorityCut depends only on an occlusion mask and a background mask and does not depend on specific model architectures.
We anticipate any warp-based image animation approaches can adopt PriorityCut with proper modifications.

\paragraph{Potential applications of PriorityCut}

Given that the only dependencies of PriorityCut are an occlusion mask and a background mask,
we expect PriorityCut to be widely applicable to any research areas involving image warping, occlusion, motion or optical flow estimation such as facial expression and body pose manipulation, image inpainting, and video frame interpolation.

\section{Conclusion}
\label{sec:conclusion}

We proposed PriorityCut, a novel augmentation approach that leverages the domain knowledge to guide the model learning the spatial relationships between pixels.
PriorityCut outperforms state-of-the-art image animation models in terms of the pixel-wise difference, low-level similarity, keypoint distance, and feature embedding distance.
Our experimental results demonstrated the effectiveness of PriorityCut in minimizing the visual artifacts in state-of-the-art image animation models.


%

\bibliography{ijcai21}
\bibliographystyle{named}

\clearpage

\setcounter{table}{0}
\setcounter{figure}{0}
\renewcommand{\thetable}{A\arabic{table}}
\renewcommand{\thefigure}{A\arabic{figure}}




%

\newcommand{\fix}{\marginpar{FIX}}
\newcommand{\new}{\marginpar{NEW}}


\appendix

\section{Architectures and Training Details}
\label{sec:appx_expr}

\paragraph{Architectures}
We followed~\citet{schonfeld2020u} to adapt a U-Net architecture in the discriminator.
A U-Net discriminator~$D^U$ consists of both an encoder~$D^U_{enc}$ and a decoder~$D^U_{dec}$.
The U-Net~encoder~$D^U_{enc}$ progressively downsamples the input image to predict global realism like the discriminator of a vanilla GAN.
The U-Net decoder~$D^U_{dec}$ progressively upsamples the feature map to predict the realism of individual pixels.
It uses skip connections to feed information between matching resolutions of $D^U_{enc}$~and~$D^U_{dec}$, enhancing its capability in segmenting fine details.
The U-Net discriminator learns both the global and the local differences between real and fake images.

Table~\ref{tab:discriminator} summarizes the differences between our discriminator architecture and that of First Order Motion Model+\footnote{\url{https://github.com/AliaksandrSiarohin/first-order-model}},
the concurrent work of First Order Motion Model~\citep{siarohin2019first} with a global discriminator.
Our U-Net encoder uses the same architecture like that of the global discriminator of First Order Motion Model+.
Our U-Net decoder is a symmetry of the U-Net encoder architecture, with skip connections feeding information between the matching resolutions of the two modules.
Both discriminators use $ch = 64$ for channels.
For the keypoint detector and the generator, we use the same architectures as those of the baseline First Order Motion Model.

\begin{table*}[!t]
	\begin{subtable}[t]{.5\linewidth}
		\centering
		\def\arraystretch{1.5}
		\caption{First Order Motion Model+ Discriminator}
		\begin{tabular}{c}
			\hline
			\hline
			RGB image $x \in \mathds{R}^{256 \times 256 \times 3}$ \\
			\hline
			ConvBlock down $ch \rightarrow 2ch$                    \\
			\hline
			ConvBlock down $2ch \rightarrow 4ch$                   \\
			\hline
			ConvBlock down $4h \rightarrow 8ch$                    \\
			\hline
			Conv2d $8ch \rightarrow 1$                             \\
			\hline
			\hline
		\end{tabular}
	\end{subtable}%
	\begin{subtable}[t]{.5\linewidth}
		\centering
		\def\arraystretch{1.5}
		\caption{U-Net Discriminator}
		\begin{tabular}{c}
			\hline
			\hline
			RGB image $x \in \mathds{R}^{256 \times 256 \times 3}$ \\
			\hline
			ConvBlock down $ch \rightarrow 2ch$                    \\
			\hline
			ConvBlock down $2ch \rightarrow 4ch$                   \\
			\hline
			ConvBlock down $4h \rightarrow 8ch$ *(see below)       \\
			\hline
			\hline
			ConvBlock up $8ch \rightarrow 4ch$                     \\
			\hline
			ConvBlock up $(4 + 4)ch \rightarrow 2ch$               \\
			\hline
			ConvBlock up $(2 + 2)ch \rightarrow ch$                \\
			\hline
			ConvBlock up $(ch + ch) \rightarrow ch$                \\
			\hline
			Conv2d $ch \rightarrow 1$                              \\
			\hline
			Sigmoid                                                \\
			\hline
			\hline
			*Conv2d $8ch \rightarrow 1$                            \\
			\hline
			\hline
		\end{tabular}
	\end{subtable}

	\caption{The architectures of First~Order~Motion Model+ discriminator and our U-Net discriminator.}
	\label{tab:discriminator}
\end{table*}

\paragraph{Hyperparameters}
We followed~\citet{siarohin2019first} to use the Adam~\citep{kingma2014adam} optimizer for training with learning rate~2e-4 and batch~size~20.
For regularization with PriorityCut, we followed~\citet{schonfeld2020u} to linearly increase the probability of augmentation from 0~to~0.5 for the first $n$ epochs.
This gives the generator sufficient time to warm up and does not make the discriminator too strong in the beginning.

\paragraph{Training losses}
Following~\citet{siarohin2019first} and~\citet{schonfeld2020u}, we used a combination of losses for both the discriminator and the generator.
The loss of the U-Net discriminator~$D^U$ consists of the adversarial loss of the U-Net encoder~$\mathcal{L}_{D_{enc}^U}$,
the adversarial~loss of the U-Net~decoder~$\mathcal{L}_{D_{dec}^U}$,
and the consistency regularization loss~$\mathcal{L}_{D_{dec}^U}^{cons}$ for the CutMix operation between the real and the fake images.
\begin{equation}
	\mathcal{L}_{D^U} = \mathcal{L}_{D_{enc}^U} +  \mathcal{L}_{D_{dec}^U} + \lambda \mathcal{L}_{D_{dec}^U}^{cons}
\end{equation}
In particular, the consistency regularization loss consists of two terms.
Suppose $\mathcal{M}_{pc}$ is the PriorityCut mask.
The first term is the output of the U-Net decoder~$D_{dec}^U$ on the CutMix image.
The second term is the CutMix between the outputs of $D_{dec}^U$ on real and fake images.
The consistency regularization loss computes the $L^2$ norm between the first and the second terms.
\begin{equation}
	\begin{split}
		\mathcal{L}_{D_{dec}^U}^{cons} = \| D_{dec}^U & \left( \text{mix}(x, x', \mathcal{M}_{pc}) \right) - \\
		\text{mix} & \left( D_{dec}^U(x), D_{dec}^U(x'), \mathcal{M}_{pc} \right) \|^2
	\end{split}
\end{equation}

The loss of the generator consists of the reconstruction~loss~$\mathcal{L}_{rec}$,
the equivariance~loss~$\mathcal{L}_{equiv}$,
the adversarial~loss~$\mathcal{L}_{adv}$,
and the feature~matching~loss~$\mathcal{L}_{feat}$.
The published First Order Motion Model paper uses only the reconstruction loss~$\mathcal{L}_{rec}$ and the equivariance~loss~$\mathcal{L}_{equiv}$~\citep{siarohin2019first}.
We followed its concurrent work to include the adversarial~loss~$\mathcal{L}_{adv}$ and the feature~matching~loss~$\mathcal{L}_{feat}$ for adversarial training.
\begin{equation}
	\mathcal{L}_{G} = \mathcal{L}_{rec} + \mathcal{L}_{equiv} + \mathcal{L}_{adv} + \mathcal{L}_{feat}
\end{equation}
The reconstruction loss~$\mathcal{L}_{rec}$ is the multi-scale perceptual loss based on the activations of the pre-trained VGG-19 network~\citep{simonyan2014very} between real and fake images.
The equivariance loss~$\mathcal{L}_{equiv}$ encourages consistent keypoint predictions given known geometric transformations.
It is based on the following equivariance constraint on local motion approximations:
\begin{equation}
	\mathcal{T}_{\textbf{X} \leftarrow \textbf{R}} \equiv \mathcal{T}_{\textbf{X} \leftarrow \textbf{Y}} \circ \mathcal{T}_{\textbf{Y} \leftarrow \textbf{R}}
\end{equation}
The adversarial loss~$\mathcal{L}_{adv}$ is the sum of the feedback from the encoder~$D_{enc}^U$ and the decoder~$D_{dec}^U$ of the U-Net discriminator~$D^U$.
The feature matching loss~$\mathcal{L}_{feat}$ matches the feature maps of each layer of the U-Net encoder~$\mathcal{L}_{D_{enc}^U}$ between the driving and the generated images,
similar to that of pix2pixHD~\citep{wang2018pix2pixHD}.

\section{Experimental Setup Details}
\label{sec:appx_eval}

\paragraph{Datasets}
We followed the preprocessing protocols of~\citet{siarohin2019first} to obtain high-quality videos on the following three datasets.
At the time of data collection, some videos are no longer available on YouTube.
We report the number of videos used in our experiments.

\begin{itemize}[leftmargin=*]
	\item The \textsf{VoxCeleb} dataset~\citep{nagrani2017voxceleb} is a face dataset with 22,496 videos from YouTube.
	      We tracked the face until it is too far from its initial position and cropped the frames using the smallest crop containing all the bounding boxes.
	      Then, we removed videos of resolution lower than $256 \times 256$ and resized the remaining videos to $256 \times 256$.
	      After preprocessing, we obtained 18,398 videos for training and 512 videos for testing.

	\item The \textsf{BAIR} robot pushing dataset~\citep{ebert2017self} contains videos of a Sawyer robotic arm pushing objects over a table.
	      It contains 42,880 videos for training and 128 videos for testing.

	\item The \textsf{Tai-Chi-HD} dataset~\citep{siarohin2019first} contains 280 tai-chi videos from YouTube.
	      We used similar preprocessing steps as \textsf{VoxCeleb} to split the videos into short clips and resized all high-quality videos to $256 \times 256$.
	      After preprocessing, we obtained 2,994 video chunks for training and 285 video chunks for testing.

\end{itemize}


Unlike First Order Motion Model paper~\citep{siarohin2019first}, we did not experiment on the UvA-NEMO dataset~\citep{dibekliouglu2012you}, since the \textsf{VoxCeleb} dataset is a more difficult one to learn.
The \textsf{VoxCeleb} dataset contains videos of different pose angles while the UvA-NEMO dataset contains only frontal faces.
Also, the UvA-NEMO dataset has a simple uniform dark background and the facial features are the only moving parts with subtle movements.

\paragraph{Metrics}
We provide additional details on the metrics and the third-party tools used for computation.

\begin{itemize}[leftmargin=*]
	\item $\mathcal{L}_1$.
	      This measures the pixel-wise differences between the generated and the ground~truth videos.

	\item \textit{Peak Signal-to-Noise Ratio (PSNR)}.
	      This measures the maximum possible power of a signal and the power of corrupting noise that affects the fidelity of its representation.
	      It evaluates how good the model is at reconstructing the low-level details of the ground~truth videos.

	\item \textit{Structural Similarity (SSIM)}.
	      This compares the low-level structures between the generated and the ground~truth videos.
	      It evaluates also how good the model is at reconstructing the low-level details.

	\item \textit{Masked PSNR and SSIM (M-PSNR, M-SSIM)}.
	      We used a variety of masks to evaluate the masked PSNR and SSIM on both the foreground and the background.
	      The salient masks in Tables~\ref{tab:sota_vox}~and~\ref{tab:sota_taichi} cover the most noticeable parts of the images.
	      The top-$k$ masks in Tables~\ref{tab:sota_vox_appx},~\ref{tab:sota_bair_appx} and~\ref{tab:sota_taichi_appx} correspond to the parts of the images that are the most difficult to inpaint.

	      To compute a salient mask, we first used DeepLabv3+~\citep{chen2018encoder}, a semantic image segmentation library, to generate a binary mask.
	      Then, we used an automatic trimap generator~\citep{lnugraha2018} to generate a trimap from the binary mask.
	      We used the image matting library $F$,~$B$,~$\alpha$~Matting~\citep{forte2020f} to generate a final alpha mask for the foreground.
	      The background mask is the inversion of the foreground mask.

	      We used the same methodology as Section~\ref{sec:priority-cut} to obtain the masks based on the top-$k$ percent occluded pixels.
	      The top-$k$ masks are derived from the occlusion masks predicted by the \textit{baseline}~model~(FOMM).
	      Similar to salient masks, we used the top-$k$ masks to evaluate the foreground and their inversions to evaluate the background.

	\item \textit{Average Keypoint Distance (AKD)}.
	      This measures the average keypoint distance between the generated and the ground~truth videos.
	      Following \citet{siarohin2019first}, we used a face alignment library~\citep{bulat2017far} to detect face keypoints for the \textsf{VoxCeleb} dataset and a human pose estimation library~\citep{cao2017realtime} to detect pose keypoints for the \textsf{Tai-Chi-HD} dataset.

	\item \textit{Missing Keypoint Rate (MKR)}.
	      This measures the percentage of keypoints detected in the ground~truth videos but not in the generated ones.
	      This evaluates the appearance quality of the generated videos.
	      Following \citet{siarohin2019first}, we used the binary label returned by the human pose estimation library~\citep{cao2017realtime} to count if a keypoint is detected or not for the \textsf{Tai-Chi-HD} dataset.

	\item \textit{Average Euclidean Distance (AED)}.
	      This measures the distance of the feature embedding between the generated and the ground~truth videos.
	      Following \citet{siarohin2019first}, we used OpenFace~\citep{baltrusaitis2018openface} to extract the face identity embedding for the \textsf{VoxCeleb} dataset and a person re-id library~\citep{hermans2017defense} to extract the person re-identification embedding for the \textsf{Tai-Chi-HD} dataset.

\end{itemize}

\begin{table*}[t]
	\begin{subtable}{\textwidth}

		\def\arraystretch{1.2}
		\setlength\tabcolsep{1.5pt}
		\centering
		\small
		\begin{tabular}{@{\extracolsep{0.005\textwidth}}*{11}{l}}
			\toprule
			$k$                                          &
			\multicolumn{2}{c}{10\%}                     & \multicolumn{2}{c}{20\%}             & \multicolumn{2}{c}{30\%}                     & \multicolumn{2}{c}{40\%}                       & \multicolumn{2}{c}{50\%}                                                                                                                                                                                                                                                                      \\
			\cmidrule{2-3} \cmidrule{4-5} \cmidrule{6-7} \cmidrule{8-9} \cmidrule{10-11}
			                                             &
			\multicolumn{1}{c}{top-$k$}                  & \multicolumn{1}{c}{$\neg$ top-$k$}   & \multicolumn{1}{c}{top-$k$}                  & \multicolumn{1}{c}{$\neg$ top-$k$}             & \multicolumn{1}{c}{top-$k$}                  & \multicolumn{1}{c}{$\neg$ top-$k$}             & \multicolumn{1}{c}{top-$k$}                  & \multicolumn{1}{c}{$\neg$ top-$k$}             & \multicolumn{1}{c}{top-$k$}                  & \multicolumn{1}{c}{$\neg$ top-$k$}             \\
			\midrule
			X2Face                                       &
			28.42\scriptsize{$\pm$0.02}                  & 19.94\scriptsize{$\pm$0.02}          & 25.41\scriptsize{$\pm$0.02}                  & 20.74\scriptsize{$\pm$0.02}                    & 23.72\scriptsize{$\pm$0.02}                  & 21.59\scriptsize{$\pm$0.02}                    & 22.57\scriptsize{$\pm$0.02}                  & 22.51\scriptsize{$\pm$0.02}                    & 21.70\scriptsize{$\pm$0.02}                  & 23.55\scriptsize{$\pm$0.02}                    \\
			Monkey-Net                                   &
			30.81\scriptsize{$\pm$0.02}                  & 23.50\scriptsize{$\pm$0.02}          & 28.00\scriptsize{$\pm$0.02}                  & 24.44\scriptsize{$\pm$0.02}                    & 26.45\scriptsize{$\pm$0.02}                  & 25.41\scriptsize{$\pm$0.02}                    & 25.41\scriptsize{$\pm$0.02}                  & 26.44\scriptsize{$\pm$0.02}                    & 24.65\scriptsize{$\pm$0.02}                  & 27.57\scriptsize{$\pm$0.02}                    \\
			FOMM                                         &
			32.99\scriptsize{$\pm$0.02}                  & 25.24\scriptsize{$\pm$0.02}          & 30.07\scriptsize{$\pm$0.02}                  & 26.14\scriptsize{$\pm$0.02}                    & 28.46\scriptsize{$\pm$0.02}                  & 27.09\scriptsize{$\pm$0.02}                    & 27.37\scriptsize{$\pm$0.02}                  & 28.11\scriptsize{$\pm$0.02}                    & 26.55\scriptsize{$\pm$0.02}                  & 29.24\scriptsize{$\pm$0.02}                    \\
			FOMM+                                        &
			\textcolor{Red}{32.91}\scriptsize{$\pm$0.02} & 25.24\scriptsize{$\pm$0.02}          & \textcolor{Red}{30.00}\scriptsize{$\pm$0.02} & \textcolor{Green}{26.16}\scriptsize{$\pm$0.02} & \textcolor{Red}{28.39}\scriptsize{$\pm$0.02} & \textcolor{Green}{27.12}\scriptsize{$\pm$0.02} & \textcolor{Red}{27.30}\scriptsize{$\pm$0.02} & \textcolor{Green}{28.15}\scriptsize{$\pm$0.02} & \textcolor{Red}{26.49}\scriptsize{$\pm$0.02} & \textcolor{Green}{29.30}\scriptsize{$\pm$0.02} \\
			Ours                                         &
			\textbf{33.14}\scriptsize{$\pm$0.02}         & \textbf{25.42}\scriptsize{$\pm$0.02} & \textbf{30.21}\scriptsize{$\pm$0.02}         & \textbf{26.34}\scriptsize{$\pm$0.02}           & \textbf{28.60}\scriptsize{$\pm$0.02}         & \textbf{27.29}\scriptsize{$\pm$0.02}           & \textbf{27.51}\scriptsize{$\pm$0.02}         & \textbf{28.32}\scriptsize{$\pm$0.02}           & \textbf{26.70}\scriptsize{$\pm$0.02}         & \textbf{29.46}\scriptsize{$\pm$0.02}           \\
			\bottomrule
		\end{tabular}
		\caption{Masked PSNR on top-$k$ percent occluded pixels.}
	\end{subtable}

	\bigskip

	\begin{subtable}{\textwidth}
		\def\arraystretch{1.2}
		\setlength\tabcolsep{1.5pt}
		\centering
		\resizebox{\textwidth}{!}{
			\begin{tabular}{@{\extracolsep{0.005\textwidth}}*{13}{l}}
				\toprule
				$k$                                           &
				\multicolumn{2}{c}{10\%}                      & \multicolumn{2}{c}{20\%}                      & \multicolumn{2}{c}{30\%}                      & \multicolumn{2}{c}{40\%}                      & \multicolumn{2}{c}{50\%}                                                                                                                                                                                                                                                                      \\
				\cmidrule{2-3} \cmidrule{4-5} \cmidrule{6-7} \cmidrule{8-9} \cmidrule{10-11}
				                                              &
				\multicolumn{1}{c}{top-$k$}                   & \multicolumn{1}{c}{$\neg$ top-$k$}            & \multicolumn{1}{c}{top-$k$}                   & \multicolumn{1}{c}{$\neg$ top-$k$}            & \multicolumn{1}{c}{top-$k$}                   & \multicolumn{1}{c}{$\neg$ top-$k$}            & \multicolumn{1}{c}{top-$k$}                   & \multicolumn{1}{c}{$\neg$ top-$k$}            & \multicolumn{1}{c}{top-$k$}                   & \multicolumn{1}{c}{$\neg$ top-$k$}            \\
				\midrule
				X2Face                                        &
				0.9519\scriptsize{$\pm$9e-5}                  & 0.6792\scriptsize{$\pm$5e-4}                  & 0.9108\scriptsize{$\pm$1e-4}                  & 0.7262\scriptsize{$\pm$5e-4}                  & 0.8721\scriptsize{$\pm$2e-4}                  & 0.7702\scriptsize{$\pm$4e-4}                  & 0.8344\scriptsize{$\pm$2e-4}                  & 0.8114\scriptsize{$\pm$4e-4}                  & 0.7977\scriptsize{$\pm$3e-4}                  & 0.8499\scriptsize{$\pm$3e-4}
				\\
				Monkey-Net                                    &
				0.9640\scriptsize{$\pm$7e-5}                  & 0.7755\scriptsize{$\pm$4e-4}                  & 0.9333\scriptsize{$\pm$1e-4}                  & 0.8134\scriptsize{$\pm$4e-4}                  & 0.9041\scriptsize{$\pm$2e-4}                  & 0.8473\scriptsize{$\pm$3e-4}                  & 0.8757\scriptsize{$\pm$2e-4}                  & 0.8779\scriptsize{$\pm$3e-4}                  & 0.8483\scriptsize{$\pm$3e-4}                  & 0.9053\scriptsize{$\pm$2e-4}
				\\
				FOMM                                          &
				0.9736\scriptsize{$\pm$6e-5}                  & 0.8270\scriptsize{$\pm$4e-4}                  & 0.9497\scriptsize{$\pm$1e-4}                  & 0.8573\scriptsize{$\pm$3e-4}                  & 0.9264\scriptsize{$\pm$2e-4}                  & 0.8842\scriptsize{$\pm$3e-4}                  & 0.9034\scriptsize{$\pm$2e-4}                  & 0.9085\scriptsize{$\pm$2e-4}                  & 0.8813\scriptsize{$\pm$3e-4}                  & 0.9301\scriptsize{$\pm$2e-4}
				\\
				FOMM+                                         &
				\textcolor{Red}{0.9734}\scriptsize{$\pm$6e-5} & \textcolor{Red}{0.8259}\scriptsize{$\pm$4e-4} & \textcolor{Red}{0.9493}\scriptsize{$\pm$1e-4} & \textcolor{Red}{0.8563}\scriptsize{$\pm$3e-4} & \textcolor{Red}{0.9260}\scriptsize{$\pm$2e-4} & \textcolor{Red}{0.8834}\scriptsize{$\pm$3e-4} & \textcolor{Red}{0.9028}\scriptsize{$\pm$2e-4} & \textcolor{Red}{0.9079}\scriptsize{$\pm$2e-4} & \textcolor{Red}{0.8805}\scriptsize{$\pm$3e-4} & \textcolor{Red}{0.9297}\scriptsize{$\pm$2e-4}
				\\
				Ours                                          &
				\textbf{0.9741}\scriptsize{$\pm$6e-5}         & \textbf{0.8287}\scriptsize{$\pm$4e-4}         & \textbf{0.9505}\scriptsize{$\pm$1e-4}         & \textbf{0.8587}\scriptsize{$\pm$1e-4}         & \textbf{0.9275}\scriptsize{$\pm$2e-4}         & \textbf{0.8854}\scriptsize{$\pm$3e-4}         & \textbf{0.9048}\scriptsize{$\pm$2e-4}         & \textbf{0.9094}\scriptsize{$\pm$2e-4}         & \textbf{0.8829}\scriptsize{$\pm$2e-4}         & \textbf{0.9308}\scriptsize{$\pm$2e-4}
				\\
				\bottomrule
			\end{tabular}
		}
		\caption{Masked SSIM on top-$k$ percent occluded pixels.}
	\end{subtable}
	\caption{Comparison with state-of-the-art approaches for video reconstruction on \textsf{VoxCeleb}. Bold indicates the best results. For variants of FOMM that do not produce the best results, red and green indicate worse and better results compared to the baseline FOMM, respectively.}
	\label{tab:sota_vox_appx}
\end{table*}

\begin{table*}[!h]
	\begin{subtable}{\textwidth}

		\def\arraystretch{1.2}
		\setlength\tabcolsep{1.5pt}
		\centering
		\small
		\begin{tabular}{@{\extracolsep{0.005\textwidth}}*{11}{l}}
			\toprule
			$k$                                 &
			\multicolumn{2}{c}{10\%}            & \multicolumn{2}{c}{20\%}            & \multicolumn{2}{c}{30\%}            & \multicolumn{2}{c}{40\%}            & \multicolumn{2}{c}{50\%}                                                                                                                                                                                                          \\
			\cmidrule{2-3} \cmidrule{4-5} \cmidrule{6-7} \cmidrule{8-9} \cmidrule{10-11}
			                                    &
			\multicolumn{1}{c}{top-$k$}         & \multicolumn{1}{c}{$\neg$ top-$k$}  & \multicolumn{1}{c}{top-$k$}         & \multicolumn{1}{c}{$\neg$ top-$k$}  & \multicolumn{1}{c}{top-$k$}         & \multicolumn{1}{c}{$\neg$ top-$k$}  & \multicolumn{1}{c}{top-$k$}         & \multicolumn{1}{c}{$\neg$ top-$k$}  & \multicolumn{1}{c}{top-$k$}         & \multicolumn{1}{c}{$\neg$ top-$k$}  \\
			\midrule
			X2Face                              &
			24.88\scriptsize{$\pm$0.1}          & 24.86\scriptsize{$\pm$0.1}          & 23.10\scriptsize{$\pm$0.1}          & 27.42\scriptsize{$\pm$0.2}          & 22.39\scriptsize{$\pm$0.1}          & 29.53\scriptsize{$\pm$0.2}          & 22.01\scriptsize{$\pm$0.1}          & 31.42\scriptsize{$\pm$0.2}          & 21.78\scriptsize{$\pm$0.1}          & 33.23\scriptsize{$\pm$0.2}          \\
			Monkey-Net                          &
			27.11\scriptsize{$\pm$0.1}          & 26.21\scriptsize{$\pm$0.1}          & 25.09\scriptsize{$\pm$0.1}          & 28.69\scriptsize{$\pm$0.2}          & 24.29\scriptsize{$\pm$0.1}          & 30.80\scriptsize{$\pm$0.2}          & 23.86\scriptsize{$\pm$0.1}          & 32.74\scriptsize{$\pm$0.1}          & 23.59\scriptsize{$\pm$0.1}          & 34.70\scriptsize{$\pm$0.1}          \\
			FOMM                                &
			29.94\scriptsize{$\pm$0.1}          & 27.24\scriptsize{$\pm$0.1}          & 27.51\scriptsize{$\pm$0.1}          & 29.40\scriptsize{$\pm$0.1}          & 26.48\scriptsize{$\pm$0.1}          & 31.38\scriptsize{$\pm$0.1}          & 25.91\scriptsize{$\pm$0.1}          & 33.27\scriptsize{$\pm$0.1}          & 25.53\scriptsize{$\pm$0.1}          & 35.21\scriptsize{$\pm$0.1}          \\
			Ours                                &
			\textbf{30.71}\scriptsize{$\pm$0.1} & \textbf{27.61}\scriptsize{$\pm$0.1} & \textbf{28.19}\scriptsize{$\pm$0.1} & \textbf{29.73}\scriptsize{$\pm$0.1} & \textbf{27.11}\scriptsize{$\pm$0.1} & \textbf{31.67}\scriptsize{$\pm$0.1} & \textbf{26.49}\scriptsize{$\pm$0.1} & \textbf{33.53}\scriptsize{$\pm$0.1} & \textbf{26.09}\scriptsize{$\pm$0.1} & \textbf{35.42}\scriptsize{$\pm$0.1} \\
			\bottomrule
		\end{tabular}
		\caption{Masked PSNR on top-$k$ percent occluded pixels.}
	\end{subtable}

	\bigskip

	\begin{subtable}{\textwidth}
		\def\arraystretch{1.2}
		\setlength\tabcolsep{1.5pt}
		\centering
		\resizebox{\textwidth}{!}{
			\begin{tabular}{@{\extracolsep{0.005\textwidth}}*{13}{l}}
				\toprule
				$k$                                   &
				\multicolumn{2}{c}{10\%}              & \multicolumn{2}{c}{20\%}              & \multicolumn{2}{c}{30\%}              & \multicolumn{2}{c}{40\%}              & \multicolumn{2}{c}{50\%}                                                                                                                                                                                                                      \\
				\cmidrule{2-3} \cmidrule{4-5} \cmidrule{6-7} \cmidrule{8-9} \cmidrule{10-11}
				                                      &
				\multicolumn{1}{c}{top-$k$}           & \multicolumn{1}{c}{$\neg$ top-$k$}    & \multicolumn{1}{c}{top-$k$}           & \multicolumn{1}{c}{$\neg$ top-$k$}    & \multicolumn{1}{c}{top-$k$}           & \multicolumn{1}{c}{$\neg$ top-$k$}    & \multicolumn{1}{c}{top-$k$}           & \multicolumn{1}{c}{$\neg$ top-$k$}    & \multicolumn{1}{c}{top-$k$}           & \multicolumn{1}{c}{$\neg$ top-$k$}    \\
				\midrule
				X2Face                                &
				0.9409\scriptsize{$\pm$4e-4}          & 0.8885\scriptsize{$\pm$2e-3}          & 0.9097\scriptsize{$\pm$9e-4}          & 0.9228\scriptsize{$\pm$1e-3}          & 0.8906\scriptsize{$\pm$1e-3}          & 0.9429\scriptsize{$\pm$1e-3}          & 0.8772\scriptsize{$\pm$1e-3}          & 0.9564\scriptsize{$\pm$8e-4}          & 0.8670\scriptsize{$\pm$2e-3}          & 0.9661\scriptsize{$\pm$7e-4}
				\\
				Monkey-Net                            &
				0.9608\scriptsize{$\pm$4e-4}          & 0.9112\scriptsize{$\pm$1e-3}          & 0.9355\scriptsize{$\pm$8e-4}          & 0.9393\scriptsize{$\pm$1e-3}          & 0.9188\scriptsize{$\pm$1e-3}          & 0.9565\scriptsize{$\pm$8e-4}          & 0.9068\scriptsize{$\pm$1e-3}          & 0.9681\scriptsize{$\pm$6e-4}          & 0.8977\scriptsize{$\pm$1e-3}          & 0.9765\scriptsize{$\pm$4e-4}
				\\
				FOMM                                  &
				0.9723\scriptsize{$\pm$4e-4}          & 0.9218\scriptsize{$\pm$1e-3}          & 0.9512\scriptsize{$\pm$7e-4}          & 0.9449\scriptsize{$\pm$9e-4}          & 0.9365\scriptsize{$\pm$9e-4}          & 0.9601\scriptsize{$\pm$7e-4}          & 0.9255\scriptsize{$\pm$1e-3}          & 0.9707\scriptsize{$\pm$5e-4}          & 0.9169\scriptsize{$\pm$5e-4}          & 0.9787\scriptsize{$\pm$4e-4}
				\\
				Ours                                  &
				\textbf{0.9750}\scriptsize{$\pm$3e-4} & \textbf{0.9244}\scriptsize{$\pm$1e-3} & \textbf{0.9551}\scriptsize{$\pm$6e-4} & \textbf{0.9463}\scriptsize{$\pm$9e-4} & \textbf{0.9409}\scriptsize{$\pm$9e-4} & \textbf{0.9609}\scriptsize{$\pm$7e-4} & \textbf{0.9304}\scriptsize{$\pm$1e-3} & \textbf{0.9714}\scriptsize{$\pm$5e-4} & \textbf{0.9221}\scriptsize{$\pm$1e-3} & \textbf{0.9791}\scriptsize{$\pm$4e-4}
				\\
				\bottomrule
			\end{tabular}
		}
		\caption{Masked SSIM on top-$k$ percent occluded pixels.}
	\end{subtable}
	\caption{Comparison with state-of-the-art approaches for video reconstruction on \textsf{BAIR}. Bold indicates the best results.}
	\label{tab:sota_bair_appx}
\end{table*}

\section{Additional Evaluations}
\label{sec:add_eval}

\paragraph{Quantitative comparison}
%

We additionally evaluated masked PSNR and SSIM on state-of-the-art approaches based on the top-$k$ percent occluded pixels.
In Tables~\ref{tab:sota_vox_appx},~\ref{tab:sota_bair_appx} and~\ref{tab:sota_taichi_appx},
the columns \textit{top-$k$} represent the masks of the $k$ percent heaviest occluded pixels.
The columns \textit{$\neg$ top-$k$} are the inversions of the top-$k$~masks.
We evaluated both masks to check if the models compromise the quality of certain parts of the image to achieve the desired performance.
For both metrics, the larger the values, the better the results.
The bold texts represent the best results.
The red and green texts represent performance loss and gain compared to the baseline model (FOMM), respectively.

We evaluated the masked versions on the \textsf{VoxCeleb}, \textsf{BAIR}, and \textsf{Tai-Chi-HD} datasets.
Table~\ref{tab:sota_vox_appx} shows the results for \textsf{VoxCeleb}.
For both PSNR and SSIM, PriorityCut outperforms state-of-the-art approaches in different thresholds~$k$.
Note that adversarial training alone does not guarantee performance gains.
For PSNR, the adversarial model~(FOMM+) compromises the quality of the hardest parts to inpaint (top-$k$ masks) and pursues the easy targets ($\neg$ top-$k$ masks).
For SSIM, adversarial training performs worse in every setting.
Table~\ref{tab:sota_bair_appx} shows the results for \textsf{BAIR}.
Since the pre-trained model of FOMM+ for \textsf{BAIR} is not publicly available at the time of writing,
we evaluated \textsf{BAIR} only on the baseline FOMM.
For both PSNR and SSIM, PriorityCut consistently outperforms state-of-the-art approaches.
Table~\ref{tab:sota_taichi_appx} shows the results for \textsf{Tai-Chi-HD}.
For PSNR, PriorityCut outperforms state-of-the-art approaches in all settings.
For SSIM, PriorityCut is on par with the adversarial model.

\begin{table*}[!tbp]
	\begin{subtable}{\textwidth}

		\def\arraystretch{1.2}
		\setlength\tabcolsep{1.5pt}
		\centering
		\small
		\begin{tabular}{@{\extracolsep{0.005\textwidth}}*{11}{l}}
			\toprule
			$k$                                            &
			\multicolumn{2}{c}{10\%}                       & \multicolumn{2}{c}{20\%}                       & \multicolumn{2}{c}{30\%}                       & \multicolumn{2}{c}{40\%}                       & \multicolumn{2}{c}{50\%}                                                                                                                                                                                                                                                                            \\
			\cmidrule{2-3} \cmidrule{4-5} \cmidrule{6-7} \cmidrule{8-9} \cmidrule{10-11}
			                                               &
			\multicolumn{1}{c}{top-$k$}                    & \multicolumn{1}{c}{$\neg$ top-$k$}             & \multicolumn{1}{c}{top-$k$}                    & \multicolumn{1}{c}{$\neg$ top-$k$}             & \multicolumn{1}{c}{top-$k$}                    & \multicolumn{1}{c}{$\neg$ top-$k$}             & \multicolumn{1}{c}{top-$k$}                    & \multicolumn{1}{c}{$\neg$ top-$k$}             & \multicolumn{1}{c}{top-$k$}                    & \multicolumn{1}{c}{$\neg$ top-$k$}             \\
			\midrule
			X2Face                                         &
			27.52\scriptsize{$\pm$0.03}                    & 19.02\scriptsize{$\pm$0.02}                    & 25.05\scriptsize{$\pm$0.03}                    & 19.64\scriptsize{$\pm$0.02}                    & 23.74\scriptsize{$\pm$0.03}                    & 20.16\scriptsize{$\pm$0.02}                    & 22.83\scriptsize{$\pm$0.03}                    & 20.66\scriptsize{$\pm$0.02}                    & 22.21\scriptsize{$\pm$0.03}                    & 21.11\scriptsize{$\pm$0.02}                    \\
			Monkey-Net                                     &
			28.11\scriptsize{$\pm$0.03}                    & 19.81\scriptsize{$\pm$0.03}                    & 25.68\scriptsize{$\pm$0.03}                    & 20.43\scriptsize{$\pm$0.03}                    & 24.35\scriptsize{$\pm$0.03}                    & 20.97\scriptsize{$\pm$0.03}                    & 23.43\scriptsize{$\pm$0.03}                    & 21.50\scriptsize{$\pm$0.03}                    & 22.79\scriptsize{$\pm$0.03}                    & 21.98\scriptsize{$\pm$0.03}                    \\
			FOMM                                           &
			31.50\scriptsize{$\pm$0.04}                    & 22.01\scriptsize{$\pm$0.03}                    & 28.62\scriptsize{$\pm$0.03}                    & 22.62\scriptsize{$\pm$0.03}                    & 27.05\scriptsize{$\pm$0.03}                    & 23.20\scriptsize{$\pm$0.03}                    & 25.97\scriptsize{$\pm$0.03}                    & 23.78\scriptsize{$\pm$0.03}                    & 25.21\scriptsize{$\pm$0.03}                    & 24.34\scriptsize{$\pm$0.03}                    \\
			FOMM+                                          &
			\textcolor{Green}{31.61}\scriptsize{$\pm$0.04} & \textcolor{Green}{22.08}\scriptsize{$\pm$0.03} & \textcolor{Green}{28.68}\scriptsize{$\pm$0.03} & \textcolor{Green}{22.70}\scriptsize{$\pm$0.03} & \textcolor{Green}{27.09}\scriptsize{$\pm$0.03} & \textcolor{Green}{23.29}\scriptsize{$\pm$0.03} & \textcolor{Green}{26.00}\scriptsize{$\pm$0.03} & \textcolor{Green}{23.88}\scriptsize{$\pm$0.03} & \textcolor{Green}{25.24}\scriptsize{$\pm$0.03} & \textcolor{Green}{24.45}\scriptsize{$\pm$0.03}
			\\
			Ours                                           &
			\textbf{31.76}\scriptsize{$\pm$0.04}           & \textbf{22.26}\scriptsize{$\pm$0.03}           & \textbf{28.86}\scriptsize{$\pm$0.03}           & \textbf{22.88}\scriptsize{$\pm$0.03}           & \textbf{27.28}\scriptsize{$\pm$0.03}           & \textbf{23.47}\scriptsize{$\pm$0.03}           & \textbf{26.18}\scriptsize{$\pm$0.03}           & \textbf{24.06}\scriptsize{$\pm$0.03}           & \textbf{25.42}\scriptsize{$\pm$0.03}           & \textbf{24.63}\scriptsize{$\pm$0.03}
			\\
			\bottomrule
		\end{tabular}
		\caption{Masked PSNR on top-$k$ percent occluded pixels.}
	\end{subtable}

	\bigskip

	\begin{subtable}{\textwidth}
		\def\arraystretch{1.2}
		\setlength\tabcolsep{1.5pt}
		\centering
		\resizebox{\textwidth}{!}{
			\begin{tabular}{@{\extracolsep{0.005\textwidth}}*{13}{l}}
				\toprule
				$k$                                             &
				\multicolumn{2}{c}{10\%}                        & \multicolumn{2}{c}{20\%}                        & \multicolumn{2}{c}{30\%}                        & \multicolumn{2}{c}{40\%}                        & \multicolumn{2}{c}{50\%}                                                                                                                                                                                                                                                                        \\
				\cmidrule{2-3} \cmidrule{4-5} \cmidrule{6-7} \cmidrule{8-9} \cmidrule{10-11}
				                                                &
				\multicolumn{1}{c}{top-$k$}                     & \multicolumn{1}{c}{$\neg$ top-$k$}              & \multicolumn{1}{c}{top-$k$}                     & \multicolumn{1}{c}{$\neg$ top-$k$}              & \multicolumn{1}{c}{top-$k$}                     & \multicolumn{1}{c}{$\neg$ top-$k$}              & \multicolumn{1}{c}{top-$k$}                     & \multicolumn{1}{c}{$\neg$ top-$k$}              & \multicolumn{1}{c}{top-$k$}           & \multicolumn{1}{c}{$\neg$ top-$k$}              \\
				\midrule
				X2Face                                          &
				0.9477\scriptsize{$\pm$2e-4}                    & 0.6439\scriptsize{$\pm$1e-3}                    & 0.9069\scriptsize{$\pm$3e-4}                    & 0.6928\scriptsize{$\pm$1e-3}                    & 0.8701\scriptsize{$\pm$5e-4}                    & 0.7351\scriptsize{$\pm$1e-3}                    & 0.8348\scriptsize{$\pm$6e-4}                    & 0.7736\scriptsize{$\pm$9e-4}                    & 0.8016\scriptsize{$\pm$8e-4}          & 0.8081\scriptsize{$\pm$7e-4}
				\\
				Monkey-Net                                      &
				0.9505\scriptsize{$\pm$2e-4}                    & 0.6613\scriptsize{$\pm$1e-3}                    & 0.9116\scriptsize{$\pm$4e-4}                    & 0.7077\scriptsize{$\pm$1e-3}                    & 0.8761\scriptsize{$\pm$5e-4}                    & 0.7478\scriptsize{$\pm$1e-3}                    & 0.8422\scriptsize{$\pm$7e-4}                    & 0.7846\scriptsize{$\pm$9e-4}                    & 0.8102\scriptsize{$\pm$9e-4}          & 0.8177\scriptsize{$\pm$7e-4}
				\\
				FOMM                                            &
				0.9626\scriptsize{$\pm$2e-4}                    & 0.7034\scriptsize{$\pm$1e-3}                    & 0.9301\scriptsize{$\pm$3e-4}                    & 0.7441\scriptsize{$\pm$1e-3}                    & 0.8992\scriptsize{$\pm$5e-4}                    & 0.7803\scriptsize{$\pm$1e-3}                    & 0.8687\scriptsize{$\pm$7e-4}                    & 0.8140\scriptsize{$\pm$8e-4}                    & 0.8393\scriptsize{$\pm$8e-4}          & 0.8447\scriptsize{$\pm$7e-4}
				\\
				FOMM+                                           &
				\textbf{0.9633}\scriptsize{$\pm$2e-4}           & \textbf{0.7055}\scriptsize{$\pm$1e-3}           & \textbf{0.9310}\scriptsize{$\pm$3e-4}           & \textbf{0.7459}\scriptsize{$\pm$1e-3}           & \textbf{0.9002}\scriptsize{$\pm$5e-4}           & \textbf{0.7822}\scriptsize{$\pm$1e-3}           & \textbf{0.8697}\scriptsize{$\pm$7e-4}           & \textcolor{Green}{0.8159}\scriptsize{$\pm$8e-4} & \textbf{0.8404}\scriptsize{$\pm$8e-4} & \textcolor{Green}{0.8465}\scriptsize{$\pm$7e-4}
				\\
				Ours                                            &
				\textcolor{Green}{0.9631}\scriptsize{$\pm$2e-4} & \textcolor{Green}{0.7048}\scriptsize{$\pm$1e-3} & \textcolor{Green}{0.9307}\scriptsize{$\pm$3e-4} & \textcolor{Green}{0.7456}\scriptsize{$\pm$1e-3} & \textcolor{Green}{0.8997}\scriptsize{$\pm$5e-4} & \textcolor{Green}{0.7820}\scriptsize{$\pm$1e-3} & \textcolor{Green}{0.8690}\scriptsize{$\pm$7e-4} & \textbf{0.8160}\scriptsize{$\pm$8e-4}           & 0.8393\scriptsize{$\pm$8e-4}          & \textbf{0.8470}\scriptsize{$\pm$7e-4}

				\\
				\bottomrule
			\end{tabular}
		}
		\caption{Masked SSIM on top-$k$ percent occluded pixels.}
	\end{subtable}
	\caption{Comparison with state-of-the-art approaches for video reconstruction on \textsf{Tai-Chi-HD}. Bold indicates the best results. For variants of FOMM that do not produce the best results, green indicates better results compared to the baseline FOMM.}
	\label{tab:sota_taichi_appx}
\end{table*}

\begin{figure*}[t]
	\small
	\setlength\tabcolsep{2pt}
	\begin{tabular}{*8{>{\centering\arraybackslash} m{0.12\textwidth}}}
		Source                                                                          & Driving 1 & Driving 2 & Driving 3 & Source & Driving 1 & Driving 2 & Driving 3 \\
		\includegraphics[width=0.12\textwidth]{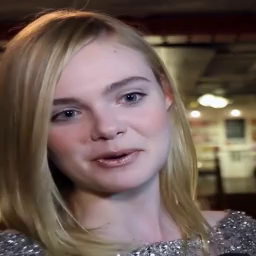}                &
		\includegraphics[width=0.12\textwidth]{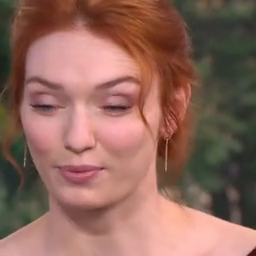}         &
		\includegraphics[width=0.12\textwidth]{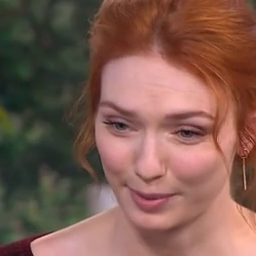}         &
		\includegraphics[width=0.12\textwidth]{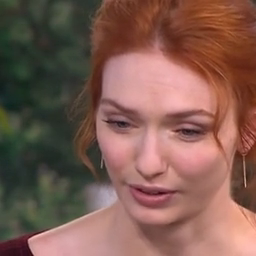}         &
		\includegraphics[width=0.12\textwidth]{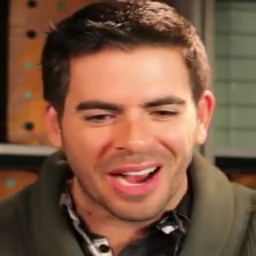}                &
		\includegraphics[width=0.12\textwidth]{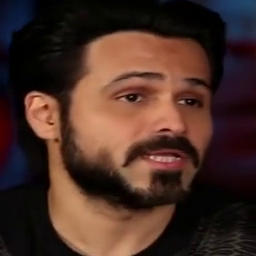}         &
		\includegraphics[width=0.12\textwidth]{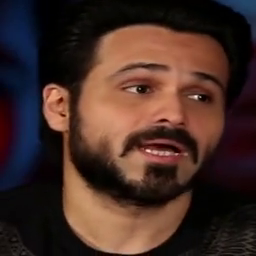}         &
		\includegraphics[width=0.12\textwidth]{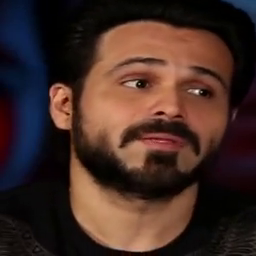}                                                                                          \\
		X2Face                                                                          &
		\includegraphics[width=0.12\textwidth]{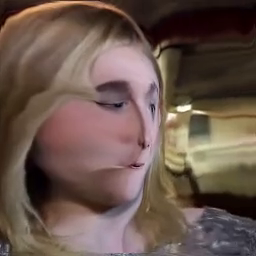}   &
		\includegraphics[width=0.12\textwidth]{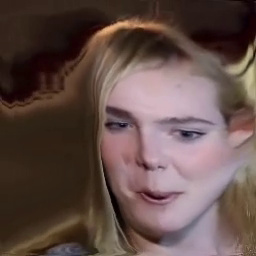}   &
		\includegraphics[width=0.12\textwidth]{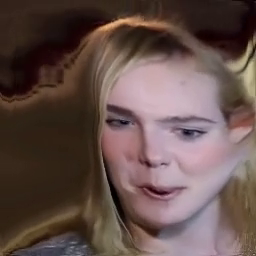}   &
		X2Face                                                                          &
		\includegraphics[width=0.12\textwidth]{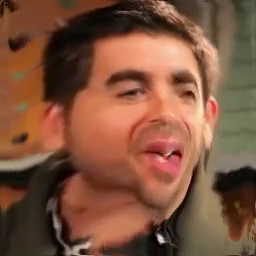}   &
		\includegraphics[width=0.12\textwidth]{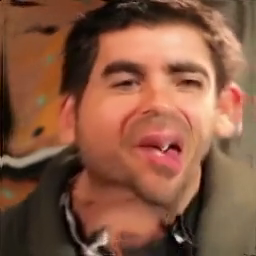}   &
		\includegraphics[width=0.12\textwidth]{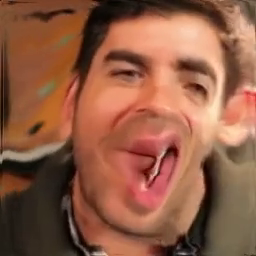}
		\\
		\makecell{Monkey-                                                                                                                                                \\ Net}                                                                       &
		\includegraphics[width=0.12\textwidth]{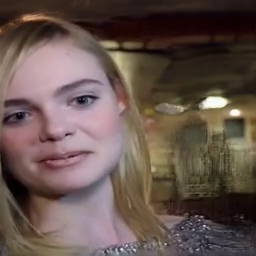}   &
		\includegraphics[width=0.12\textwidth]{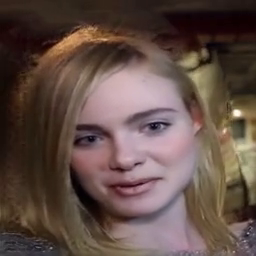}   &
		\includegraphics[width=0.12\textwidth]{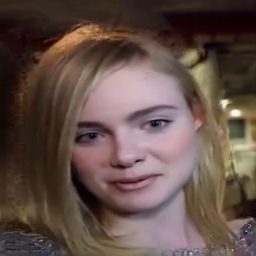}   &
		\makecell{Monkey-                                                                                                                                                \\ Net} &
		\includegraphics[width=0.12\textwidth]{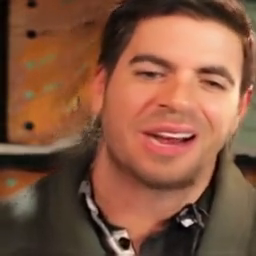}   &
		\includegraphics[width=0.12\textwidth]{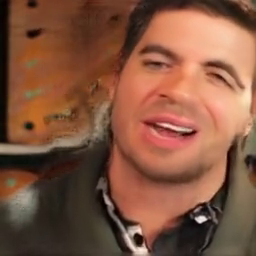}   &
		\includegraphics[width=0.12\textwidth]{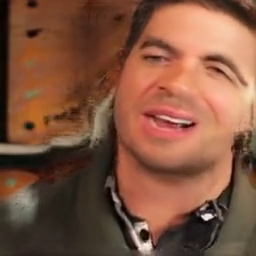}
		\\
		FOMM                                                                            &
		\includegraphics[width=0.12\textwidth]{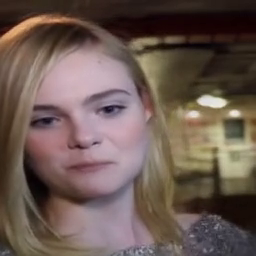} &
		\includegraphics[width=0.12\textwidth]{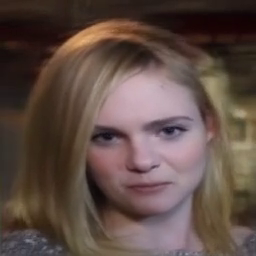} &
		\includegraphics[width=0.12\textwidth]{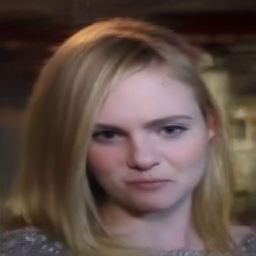} &
		FOMM                                                                            &
		\includegraphics[width=0.12\textwidth]{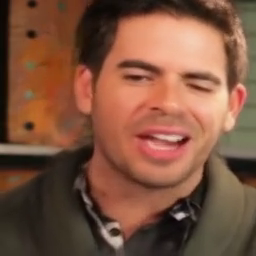} &
		\includegraphics[width=0.12\textwidth]{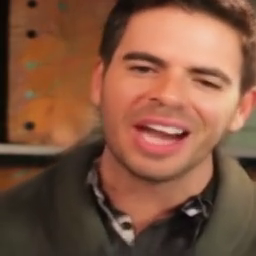} &
		\includegraphics[width=0.12\textwidth]{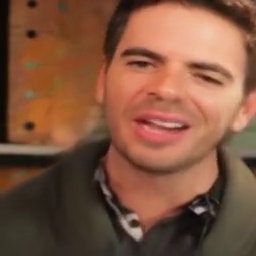}
		\\
		FOMM+                                                                           &
		\includegraphics[width=0.12\textwidth]{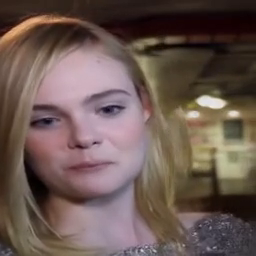}      &
		\includegraphics[width=0.12\textwidth]{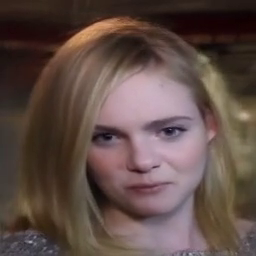}      &
		\includegraphics[width=0.12\textwidth]{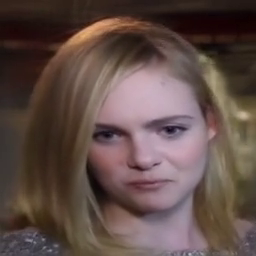}      &
		FOMM+                                                                           &
		\includegraphics[width=0.12\textwidth]{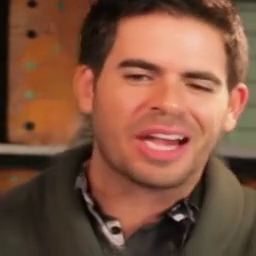}      &
		\includegraphics[width=0.12\textwidth]{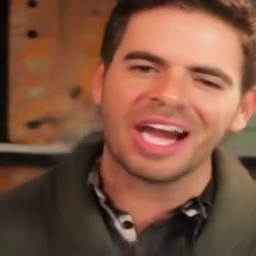}      &
		\includegraphics[width=0.12\textwidth]{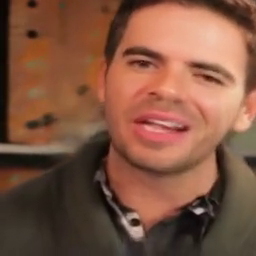}
		\\
		Ours                                                                            &
		\includegraphics[width=0.12\textwidth]{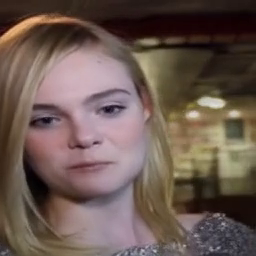} &
		\includegraphics[width=0.12\textwidth]{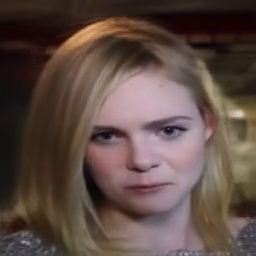} &
		\includegraphics[width=0.12\textwidth]{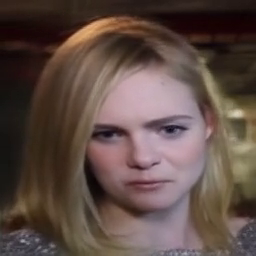} &
		Ours                                                                            &
		\includegraphics[width=0.12\textwidth]{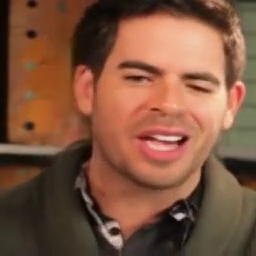} &
		\includegraphics[width=0.12\textwidth]{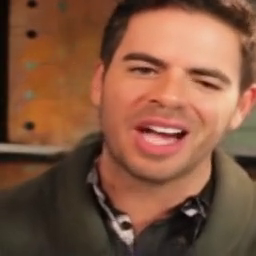} &
		\includegraphics[width=0.12\textwidth]{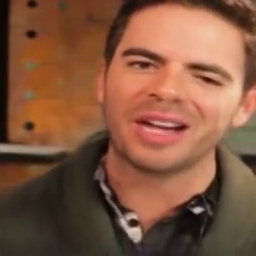}
		\\
	\end{tabular}

	\caption{Qualitative comparison of state-of-the-art approaches for image animation on \textsf{VoxCeleb}. FOMM and FOMM+ produce noticeable visual artifacts such as overlapping hairs and blurry background in the first example and confusing background texture in the second example. PriorityCut helps reduce these ambiguous visual artifacts in FOMM.}
	\label{fig:appx_vox}
\end{figure*}

\begin{figure*}[t]
	\small
	\setlength\tabcolsep{2pt}
	\begin{tabular}{*8{>{\centering\arraybackslash} m{0.12\textwidth}}}
		Source                                                                           & Driving 1 & Driving 2 & Driving 3 & Source & Driving 1 & Driving 2 & Driving 3 \\
		\includegraphics[width=0.12\textwidth]{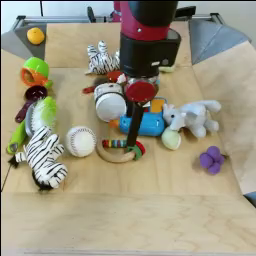}                &
		\includegraphics[width=0.12\textwidth]{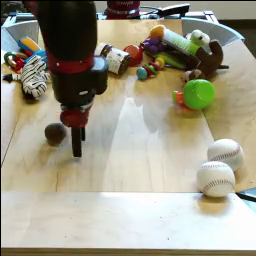}         &
		\includegraphics[width=0.12\textwidth]{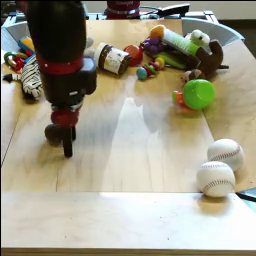}         &
		\includegraphics[width=0.12\textwidth]{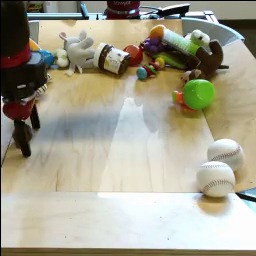}         &
		\includegraphics[width=0.12\textwidth]{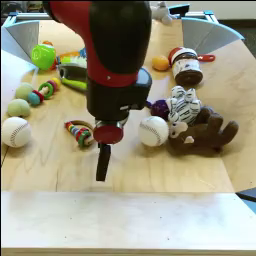}                &
		\includegraphics[width=0.12\textwidth]{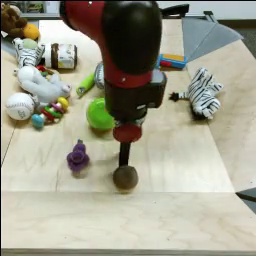}         &
		\includegraphics[width=0.12\textwidth]{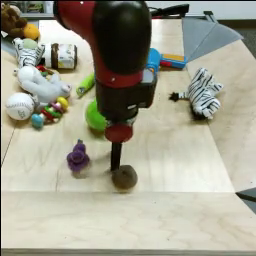}         &
		\includegraphics[width=0.12\textwidth]{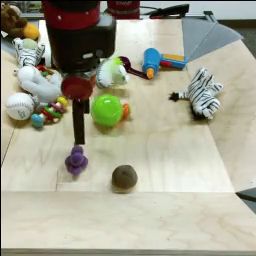}                                                                                          \\
		X2Face                                                                           &
		\includegraphics[width=0.12\textwidth]{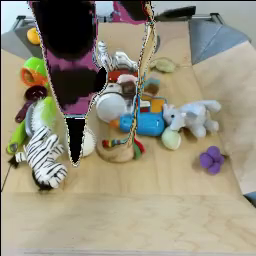}   &
		\includegraphics[width=0.12\textwidth]{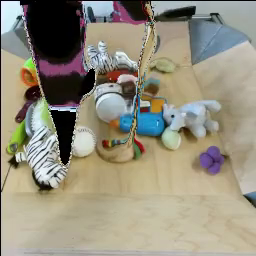}   &
		\includegraphics[width=0.12\textwidth]{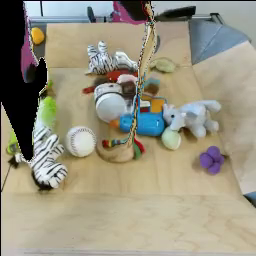}   &
		X2Face                                                                           &
		\includegraphics[width=0.12\textwidth]{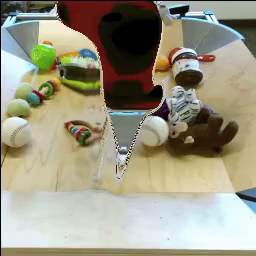}   &
		\includegraphics[width=0.12\textwidth]{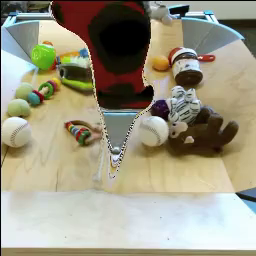}   &
		\includegraphics[width=0.12\textwidth]{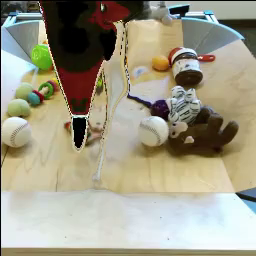}
		\\
		\makecell{Monkey-                                                                                                                                                 \\ Net}                                                                       &
		\includegraphics[width=0.12\textwidth]{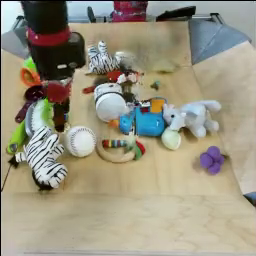}   &
		\includegraphics[width=0.12\textwidth]{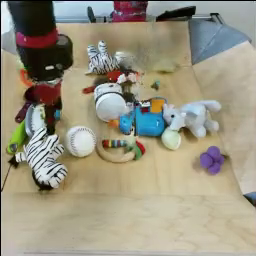}   &
		\includegraphics[width=0.12\textwidth]{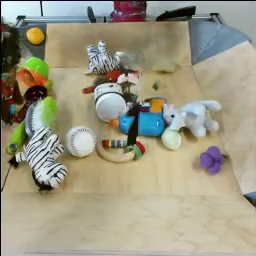}   &
		\makecell{Monkey-                                                                                                                                                 \\ Net}                                                                             &
		\includegraphics[width=0.12\textwidth]{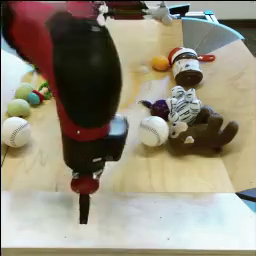}   &
		\includegraphics[width=0.12\textwidth]{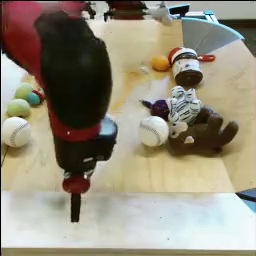}   &
		\includegraphics[width=0.12\textwidth]{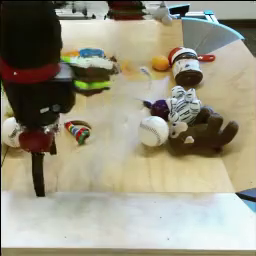}
		\\
		FOMM                                                                             &
		\includegraphics[width=0.12\textwidth]{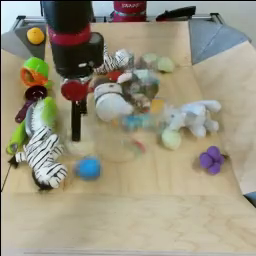} &
		\includegraphics[width=0.12\textwidth]{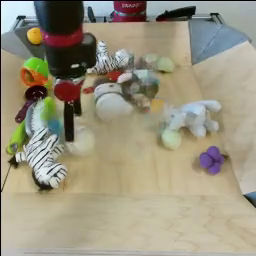} &
		\includegraphics[width=0.12\textwidth]{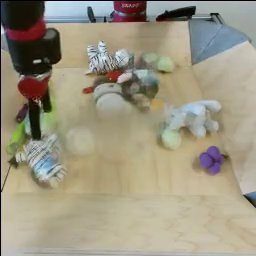} &
		FOMM                                                                             &
		\includegraphics[width=0.12\textwidth]{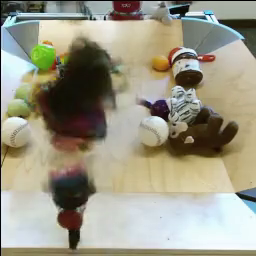} &
		\includegraphics[width=0.12\textwidth]{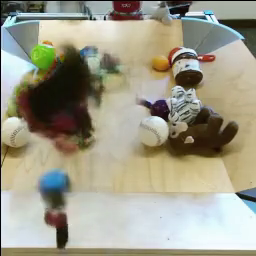} &
		\includegraphics[width=0.12\textwidth]{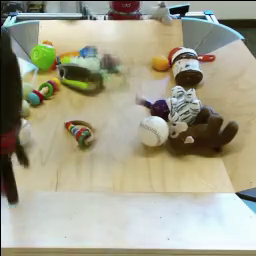}
		\\
		Ours                                                                             &
		\includegraphics[width=0.12\textwidth]{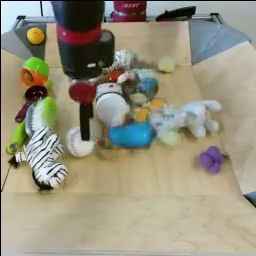} &
		\includegraphics[width=0.12\textwidth]{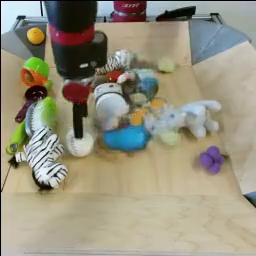} &
		\includegraphics[width=0.12\textwidth]{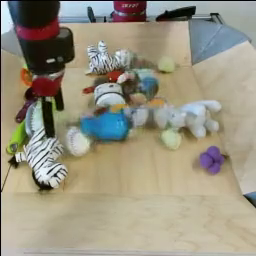} &
		FOMM                                                                             &
		\includegraphics[width=0.12\textwidth]{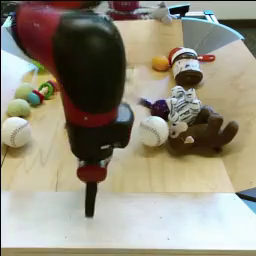} &
		\includegraphics[width=0.12\textwidth]{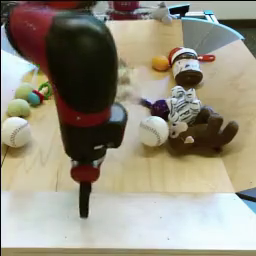} &
		\includegraphics[width=0.12\textwidth]{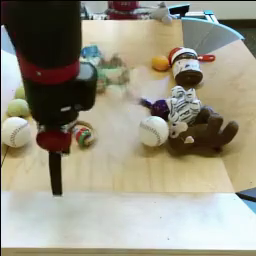}
		\\
	\end{tabular}

	\caption{Qualitative comparison of state-of-the-art approaches for image animation on \textsf{BAIR}. FOMM replaces the background objects by the background texture in the first example and produces broken texture in the second example. PriorityCut minimizes these ambiguous visual artifacts in FOMM.}
	\label{fig:appx_bair}
\end{figure*}

\begin{figure*}[t]
	\small
	\setlength\tabcolsep{2pt}
	\begin{tabular}{*8{>{\centering\arraybackslash} m{0.12\textwidth}}}
		Source                                                                             & Driving 1 & Driving 2 & Driving 3 & Source & Driving 1 & Driving 2 & Driving 3 \\
		\includegraphics[width=0.12\textwidth]{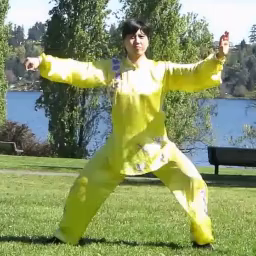}                &
		\includegraphics[width=0.12\textwidth]{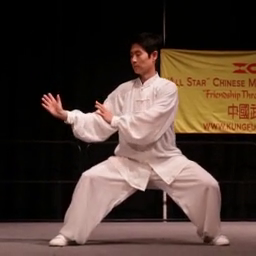}         &
		\includegraphics[width=0.12\textwidth]{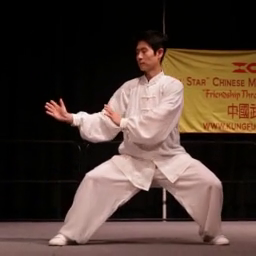}         &
		\includegraphics[width=0.12\textwidth]{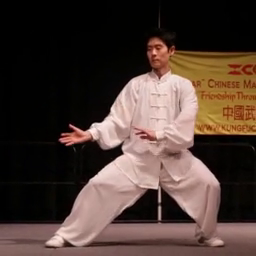}         &
		\includegraphics[width=0.12\textwidth]{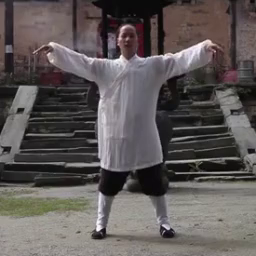}                &
		\includegraphics[width=0.12\textwidth]{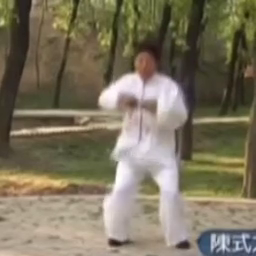}         &
		\includegraphics[width=0.12\textwidth]{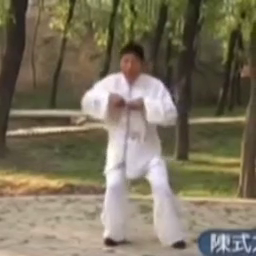}         &
		\includegraphics[width=0.12\textwidth]{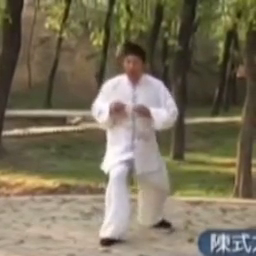}                                                                                          \\
		X2Face                                                                             &
		\includegraphics[width=0.12\textwidth]{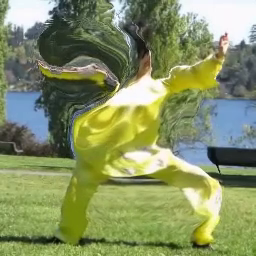}   &
		\includegraphics[width=0.12\textwidth]{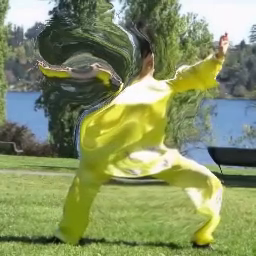}   &
		\includegraphics[width=0.12\textwidth]{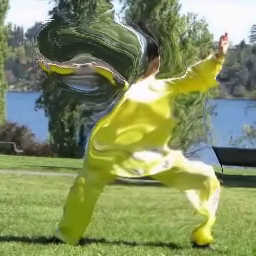}   &
		X2Face                                                                             &
		\includegraphics[width=0.12\textwidth]{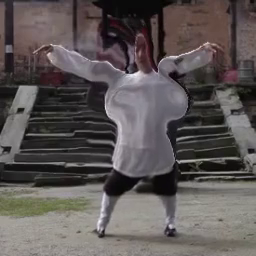}   &
		\includegraphics[width=0.12\textwidth]{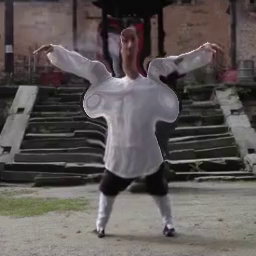}   &
		\includegraphics[width=0.12\textwidth]{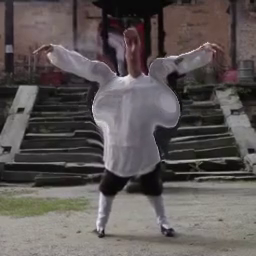}
		\\
		\makecell{Monkey-                                                                                                                                                   \\ Net}                                                                       &
		\includegraphics[width=0.12\textwidth]{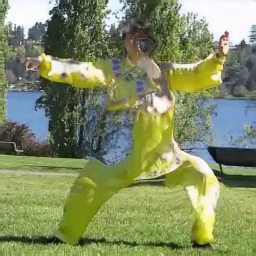}   &
		\includegraphics[width=0.12\textwidth]{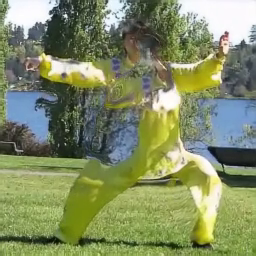}   &
		\includegraphics[width=0.12\textwidth]{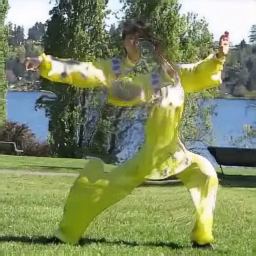}   &
		\makecell{Monkey-                                                                                                                                                   \\ Net} &
		\includegraphics[width=0.12\textwidth]{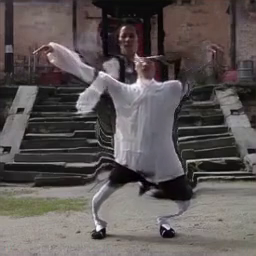}   &
		\includegraphics[width=0.12\textwidth]{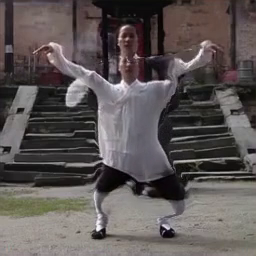}   &
		\includegraphics[width=0.12\textwidth]{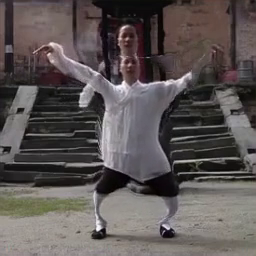}
		\\
		FOMM                                                                               &
		\includegraphics[width=0.12\textwidth]{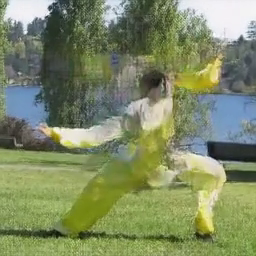} &
		\includegraphics[width=0.12\textwidth]{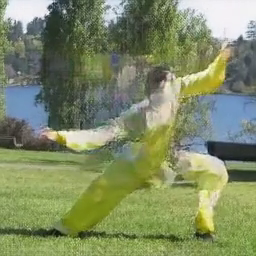} &
		\includegraphics[width=0.12\textwidth]{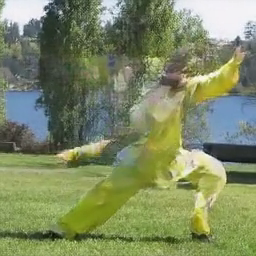} &
		FOMM                                                                               &
		\includegraphics[width=0.12\textwidth]{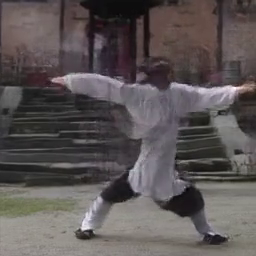} &
		\includegraphics[width=0.12\textwidth]{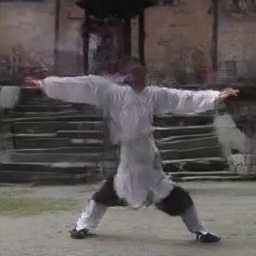} &
		\includegraphics[width=0.12\textwidth]{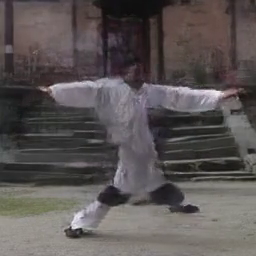}
		\\
		FOMM+                                                                              &
		\includegraphics[width=0.12\textwidth]{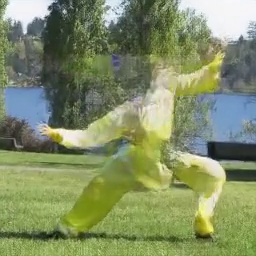}      &
		\includegraphics[width=0.12\textwidth]{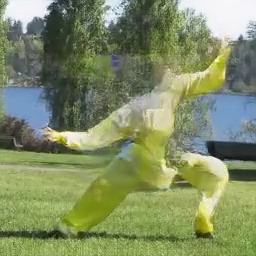}      &
		\includegraphics[width=0.12\textwidth]{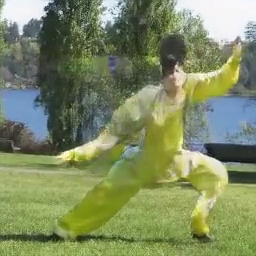}      &
		FOMM+                                                                              &
		\includegraphics[width=0.12\textwidth]{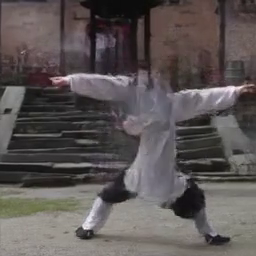}      &
		\includegraphics[width=0.12\textwidth]{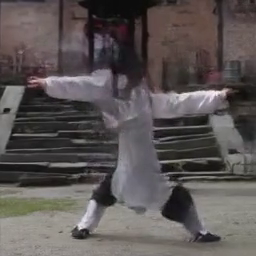}      &
		\includegraphics[width=0.12\textwidth]{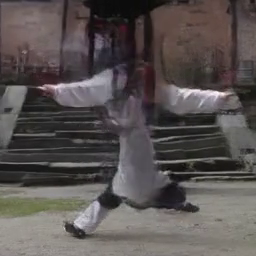}
		\\
		Ours                                                                               &
		\includegraphics[width=0.12\textwidth]{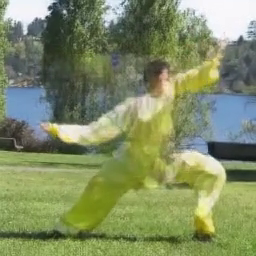} &
		\includegraphics[width=0.12\textwidth]{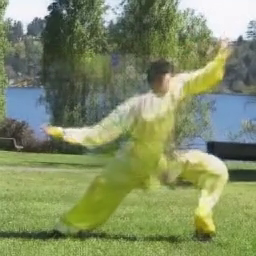} &
		\includegraphics[width=0.12\textwidth]{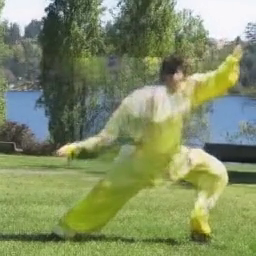} &
		Ours                                                                               &
		\includegraphics[width=0.12\textwidth]{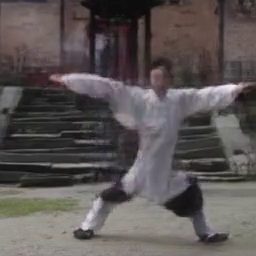} &
		\includegraphics[width=0.12\textwidth]{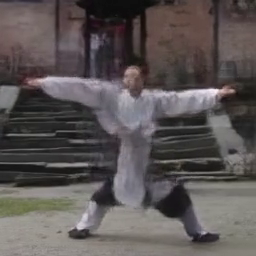} &
		\includegraphics[width=0.12\textwidth]{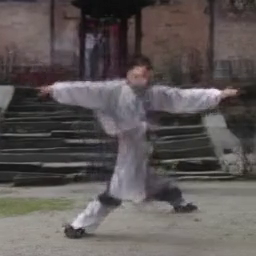}
		\\
	\end{tabular}

	\caption{Qualitative comparison of state-of-the-art approaches for image animation on \textsf{Tai-Chi-HD}. FOMM and FOMM+ inpaint the head areas with background texture while the heads are visible in the results of PriorityCut.}
	\label{fig:appx_taichi}
\end{figure*}

\paragraph{Qualitative comparison}
We performed additional qualitative comparisons between state-of-the-art approaches on each dataset.
Figures~\ref{fig:appx_vox}, ~\ref{fig:appx_bair}, and ~\ref{fig:appx_taichi} show the qualitative comparisons for \textsf{VoxCeleb}, \textsf{BAIR}, and \textsf{Tai-Chi-HD} datasets respectively.
For each figure, we include corresponding videos for reference.
Since we implemented PriorityCut on top of First Order Motion Model~(FOMM),
the artifacts seen in FOMM also exist in our results.
However, PriorityCut alleviates them by reducing the ambiguity of the baseline FOMM.

For \textsf{VoxCeleb} in Figure~\ref{fig:appx_vox},
X2Face shows severe warping artifacts and face distortions.
Monkey-Net shows noticeable warping artifacts around locations of large changes in motion.
FOMM shows overlapping hairs and blurry background in the first example and blurry texture behind the right ear in the second example.
FOMM+ further amplifies these visual artifacts in both examples.
In contrast, PriorityCut minimizes these ambiguous visual artifacts.

For \textsf{BAIR} in Figure~\ref{fig:appx_bair},
X2Face shows trivial warping artifacts and broken texture on the robot arms.
Monkey-Net shows warping artifacts around the robot arms.
FOMM erases the objects and fills with background texture in the first example and shows broken texture on the robot arms in the second example.
In contrast, PriorityCut minimizes the ambiguity without replacing the objects with background texture and preserve the robot arm texture.

For \textsf{Tai-Chi-HD} in Figure~\ref{fig:appx_taichi},
both MonkeyNet and X2Face show trivial warping artifacts.
Both FOMM and FOMM+ fill the head area with background texture while the heads are visible for PriorityCut.

Our qualitative comparisons on diverse datasets show the effectiveness of PriorityCut in alleviating the visual artifacts in state-of-the-art image animation models.

\bibliography{ijcai21}
\bibliographystyle{named}

\end{document}